\documentclass[journal]{vgtc}                     % final (journal style)
\onlineid{1254}

\vgtccategory{Research}

\vgtcpapertype{application/design study}

\title{ConceptViz: A Visual Analytics Approach for Exploring Concepts in Large Language Models}

\author{%
    Haoxuan Li, 
    Zhen Wen, 
    Qiqi Jiang, 
    Chenxiao Li, 
    Yuwei Wu, 
    Yuchen Yang, 
    Yiyao Wang, \\
    Xiuqi Huang, 
    Minfeng Zhu, 
    and Wei Chen
}

\authorfooter{
  \item
  	Haoxuan Li, Zhen Wen, Qiqi Jiang, Yuwei Wu, Yuchen Yang, Yiyao Wang, 
  	Xiuqi Huang, and Wei Chen are with the State Key Lab of CAD\&CG, Zhejiang University. 
  	E-mail: \{lihaoxuan | wenzhen | qiqijiang348284 | 22451008 | yyc\_yang | wangyiyao | chenvis\}@zju.edu.cn.
  \item
  	Chenxiao Li and Minfeng Zhu are with Zhejiang University. 
  	E-mail: \{3220101835 | minfeng\_zhu\}@zju.edu.cn.
  \item
  	Haoxuan Li and Zhen Wen contributed equally to this work.
  \item
  	Minfeng Zhu, Xiuqi Huang and Wei Chen are the corresponding authors.
}

\abstract{
Large language models (LLMs) have achieved remarkable performance across a wide range of natural language tasks. Understanding how LLMs internally represent knowledge remains a significant challenge. Despite Sparse Autoencoders (SAEs) have emerged as a promising technique for extracting interpretable features from LLMs, SAE features do not inherently align with human-understandable concepts, making their interpretation cumbersome and labor-intensive. To bridge the gap between SAE features and human concepts, we present ConceptViz, a visual analytics system designed for exploring concepts in LLMs. ConceptViz implements a novel Identification $\Rightarrow$ Interpretation $\Rightarrow$ Validation pipeline, enabling users to query SAEs using concepts of interest, interactively explore concept-to-feature alignments, and validate the correspondences through model behavior verification. We demonstrate the effectiveness of ConceptViz through two usage scenarios and a user study. Our results show that ConceptViz enhances interpretability research by streamlining the discovery and validation of meaningful concept representations in LLMs, ultimately aiding researchers in building more accurate mental models of LLM features. Our code and user guide are publicly available at \url{https://github.com/Happy-Hippo209/ConceptViz}. 
}


\teaser{
\centering
\includegraphics[width=\linewidth]{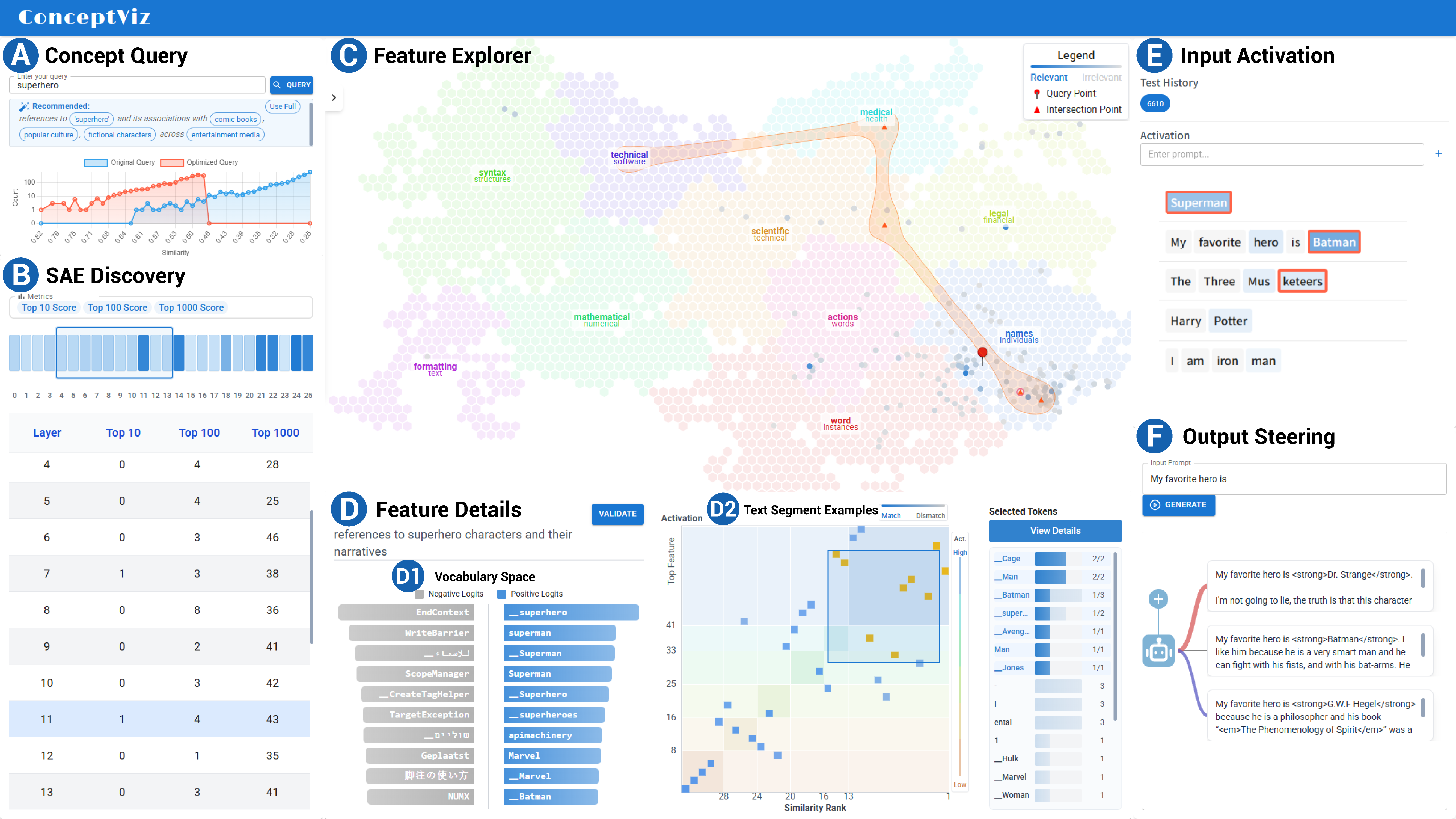}
\caption{
Our system for exploring and interpreting SAE features in LLMs consists of six interconnected views: (A) Concept Query View for building and refining queries, (B) SAE Discovery View for identifying relevant SAE models, (C) Feature Explorer View for browsing SAE features in concept space, (D) Feature Details View for examining semantic meaning of features, (E) Input Activation View for verifying response between features and concepts, and (F) Output Steering View for observing feature causal impact on model outputs.
}
\label{fig:teaser}
}

\graphicspath{{figs/}{figures/}{pictures/}{images/}{./}} % where to search for the images

\usepackage{tabu}                      % only used for the table example
\usepackage{booktabs}                  % only used for the table example
\usepackage{lipsum}                    % used to generate placeholder text
\usepackage{mwe}                       % used to generate placeholder figures

\usepackage{mathptmx}                  % use matching math font
\usepackage{amsmath}                   % for \text{} in math mode
\usepackage{pifont}

\usepackage[skins]{tcolorbox}
\usepackage[svgnames,x11names,rgb]{xcolor}

\definecolor{lightblue}{HTML}{e7f0fa}

\newcommand{\ecl}[1]{%
    \tcbox[
        on line,
        arc=0pt,
        outer arc=0pt,
        colback=lightblue,
        colframe=lightblue,
        boxsep=0pt,
        left=0pt,
        right=0pt,
        top=0pt,
        bottom=0pt
    ]{#1}%
}

\usepackage{xcolor} 
\definecolor{revisioncolor}{rgb}{1.0, 0.9, 0.9}

\ifdefined\ShowRevisions
    \newcommand{\wz}[1]{\textcolor{red}{#1}}
    \newcommand{\lhx}[1]{\textcolor{red}{#1}}
\else
    \newcommand{\wz}[1]{{#1}}
    \newcommand{\lhx}[1]{{#1}}
\fi

\usepackage[textsize=tiny]{todonotes}

\begin{document}

%%%%%%%%%%%%%%%%%%%%%%%%%%%%%%%%%%%%%%%%%%%%%%%%%%%%%%%%%%%%%%%%
%%%%%%%%%%%%%%%%%%%%%% START OF THE PAPER %%%%%%%%%%%%%%%%%%%%%%
%%%%%%%%%%%%%%%%%%%%%%%%%%%%%%%%%%%%%%%%%%%%%%%%%%%%%%%%%%%%%%%%
\firstsection{Introduction}
\maketitle

\begin{figure}[t]
\centering
\includegraphics[width=1.0\columnwidth]{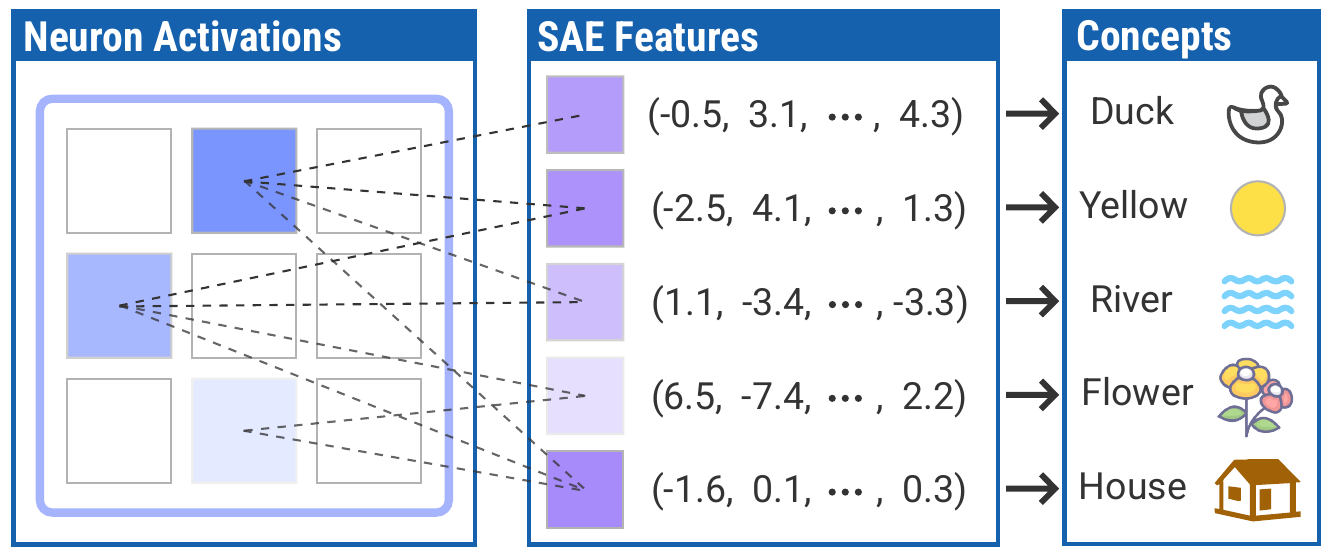}
\caption{SAEs address the polysemanticity problem in LLM neurons by projecting neuron activations into sparse, interpretable features. Each SAE feature corresponds to more specific human-understandable concepts (e.g., Duck, Yellow), enabling clearer semantic distinctions than the original neurons which may activate for multiple unrelated concepts.}
\label{fig:intro}
\vspace{-6mm}
\end{figure}

Large language models (LLMs) have demonstrated remarkable capabilities in language generation, translation, and complex reasoning through massive parameter scaling and extensive training data\cite{gemmateam2024gemma2, openai2024gpt4, TheC3}. However, their black-box nature raises interpretability concerns. Traditional interpretability methods\cite{yu-ananiadou-2024-neuron, meng2022locating, DBLP:journals/corr/abs-1906-05714, Wang2022InterpretabilityIT} focus on understanding how neural networks utilize internal knowledge and calculate their outputs. But they struggle with handling polysemanticity\cite{bricken2023monosemanticity, elhage2022superposition, olah2020zoom}, where individual neurons usually activate for multiple unrelated concepts.
Recent approaches in Sparse Autoencoders (SAEs)\lhx{\cite{shu2025surveysparseautoencodersinterpreting, bricken2023monosemanticity, cunningham2023sparseautoencodershighlyinterpretable}} address this by decomposing dense neuron activations into sparse and interpretable features (see \cref{fig:intro}). These SAE features form a concept space with clearer semantic boundaries, which reduces polysemanticity, offering new opportunities for the interpretability of LLMs.

% However, despite their potential, the application of SAEs for mechanistic interpretability remains in its early stages. A significant challenge lies in effectively evaluating the features extracted by SAEs after training. On one hand, researchers often rely on proxy metrics such as sparsity, reconstruction quality, and supervised probing tasks to indirectly assess SAE performance. Although useful, these metrics introduce an additional layer of abstraction, which can obscure subjective evaluations of interpretability. On the other hand, researchers commonly interpret the latent features of SAEs by examining the dataset samples that activate them, either through manual inspection using feature dashboards (Bricken et al., 2023) or by employing automated interpretability techniques (Gao et al., 2024). While open-source libraries like SAELens and SAEvis, along with the platform NeuronPedia, facilitate feature exploration and provide statistical summaries and basic querying tools, they require users to manually sift through vast numbers of features to identify those closely aligned with specific human concepts. This process is time-consuming, inefficient, and lacks an intuitive interactive workflow, highlighting the need for more advanced and user-centric exploration methods.

\textbf{Limitations}. However, SAE features remain difficult for humans to interpret directly, as they do not inherently align with discrete and well-defined human concepts.
Existing explanation approaches attempt to bridge this gap through quantitative metrics and qualitative descriptions, but still exhibit critical limitations. 
%proxy metric mismatch, feature discovery bottleneck, and inflexible disconnect verification.
On the one hand, proxy metrics such as sparsity, reconstruction quality, and supervised probing tasks \cite{cunningham2023sparseautoencodershighlyinterpretable, gao2024scaling} provide indirect performance assessment of SAE features, but fail to establish a clear mapping between features and human-understandable concepts.  
On the other hand, while automated interpretability techniques can leverage text segments that maximally activate specific SAE features\cite{bricken2023monosemanticity, gao2024scaling} to provide quick but non-robust \cite{durmus2024steering, wu2025interpretingsteeringllmsmutual} feature explanations, they still require manual inspection of vast numbers of features through text segments for verification, leading to cognitive overload and explanation bottlenecks.
% or apply automatic interpretability techniques\cite{bills2023language, paulo2025automatically,wu2025interpretingsteeringllmsmutual}
Although existing open-source tools\cite{bloom2024saetrainingcodebase, neuronpedia} provide dashboards for browsing features, the abstract proxy metrics or fragmented text segments scant systematic workflows to connect conceptual analysis with concrete model behaviors. 
These limitations indicate that the SAE feature explanation process is time-consuming and lacks intuitive interactivity, highlighting the need for more user-centric visual analytics systems. 
%Alternatively, researchers manually examine dataset samples that activate latent features\cite{bricken2023monosemanticity, gao2024scaling} or use automated interpretability techniques\cite{bills2023language, paulo2025automatically,wu2025interpretingsteeringllmsmutual} providing quick but non-robust\cite{durmus2024steering, wu2025interpretingsteeringllmsmutual} feature semantic overviews. Although existing open-source tools\cite{bloom2024saetrainingcodebase, neuronpedia} combine these methods, they still require users to manually sift through vast numbers of features to identify those that align with human concepts. This process is time-consuming and lacks intuitive interactivity, highlighting the need for more user-centric visual analytics (VA) methods.

%%%%%%%%%%%%%%%%%%%%%%%%%%%%%%
\textbf{Challenges}. Designing effective visual analytics systems for SAE explanation confronts three major challenges. 
\ding{172} Cross-Layer SAE Recommendation: Given the vast number of SAE models trained across different transformer layers of an LLM, selecting the most relevant model for a target concept is non-trivial. Researchers often lack systematic guidance on which SAE model best aligns with their interested concepts.
\ding{173} Multi-Scale Feature Navigation: Even within a chosen SAE model, identifying features that best correspond to a given concept remains challenging due to the high-dimensional nature of feature representations. Effective visualization techniques must maintain both the global structure of the feature space and the fine-grained insights between individual features.
\ding{174} Interactive Concept Validation: Establishing trust in selected SAE features that consistently reflect the intended concept requires thorough validation. Researchers need interactive tools that facilitate seamless transitions between high-level conceptual exploration and instance-level verification, confirming feature interpretations through steering model behaviors.
Dealing with these challenges is essential for developing effective visual analytics systems that support the interpretability of SAE features in LLMs.

\textbf{Key Technologies}. To address these challenges, we propose ConceptViz, a concept-driven visual analytics system for exploring SAE features. Our system introduces an interactive analysis pipeline that guides researchers through a three-phase workflow: Identification $\Rightarrow$ Interpretation $\Rightarrow$ Validation. First, to efficiently identify relevant SAE models \textbf{(Identification)}, we enable concept-based queries combined with structural information about model architecture, helping users discover SAE models that most strongly align with their concepts of interest. Second, to explore the alignment between feature space and human concepts in-depth \textbf{(Interpretation)}, we introduce interactive visualizations that project sparse features into a concept space, enabling detailed semantic matching and insightful feature interpretation. Third, to verify the alignment between features and human concepts at the instance level \textbf{(Validation)}, we provide input and output-based validation, ensuring that the extracted features consistently correspond to human-defined concepts throughout the LLM's processing pipeline. This integrated approach ensures a seamless transition from model discovery to detailed analysis and \wz{empirical validation}, empowering researchers to conduct comprehensive and interactive SAE exploration.

\textbf{Contributions}. In summary, the main contributions are as follows:
\begin{itemize}[leftmargin=12pt, itemsep=0pt, topsep=0mm]
\item We design an Identification--Interpretation--Validation workflow that enables concept-based SAE identification, interactive feature interpretation, and thorough instance-level validation, supporting in-depth exploration of SAE features.

\item We present ConceptViz, a novel concept-driven visual analytics system that implements this pipeline, bridging the gap between complex model representations and human-understandable concepts.

\item We evaluate ConceptViz through two usage scenarios and a user study, demonstrating its effectiveness, usability, and potential to advance interpretability research.
\end{itemize}
\vspace{-2mm}

  % 引入引言部分

\section{Related Work}

\subsection{LLM Interpretability}
%\subsection{Feature Analysis in LLMs}

Research on LLM interpretability has increasingly focused on feature analysis, which primarily involves three aspects: feature extraction, feature interpretation and evaluation, and feature application. In feature extraction, researchers initially focused on studying LLM hidden representations directly. Meng et al.\cite{meng2022locating} located neuron sets critical for factual prediction. Wang et al.\cite{Wang2022InterpretabilityIT} and Olsson et al.\cite{olsson2022context} focused on analyzing attention mechanisms. However, it is non-trivial to achieve that because of the polysemantic nature\cite{olah2020zoom, elhage2022superposition, bricken2023monosemanticity}. To address this challenge, researchers developed various decomposition techniques. Millidge et al.~\cite{beren2022singular} applied singular value decomposition to separate semantic components of neurons; Wu et al.\cite{wu-etal-2024-language} combined matrix decomposition with human instructions to improve feature interpretability. SAEs, as a more advanced technique, produce more mono-semantic feature representations by enforcing sparsity constraints that encourage selective activation. Bricken et al.\cite{bricken2023monosemanticity} and Cunningham et al.\cite{cunningham2023sparseautoencodershighlyinterpretable} demonstrated the effectiveness of SAEs on language models of different scales. Subsequent researchers proposed various SAE variants\cite{rajamanoharan2024improvingdictionarylearninggated, Rajamanoharan2024JumpingAI, gao2024scaling, karvonen2025measuringprogressdictionarylearning, shu2025inputactivationsidentifyinginfluential} to improve feature quality. Recent studies \cite{templeton2024scaling,he2024llamascopeextractingmillions} managed to extend SAE techniques to larger LLMs with tens of billions of parameters.

For feature interpretation and evaluation, researchers adopted various strategies. The most direct approach is collecting text segments that maximally activate specific features. Bricken et al.\cite{bricken2023monosemanticity} analyzed N-gram activation patterns of features across large corpora. For auto interpretation, Foote et al.\cite{foote2023ng} developed the N2G method, while Bills et al.\cite{bills2023language} and Paulo et al.~\cite{paulo2025automatically} utilized LLMs, to transform N-gram sequences into natural language explanations. Another category of evaluation methods is based on feature performance metrics such as sparsity and reconstruction error \cite{cunningham2023sparseautoencodershighlyinterpretable,gao2024scaling}. Karvonen et al.\cite{karvonen2025saebench} systematically reviewed prior evaluation methods\cite{bills2023language,marks2024sparsefeaturecircuitsdiscovering,chaudhary2024evaluatingopensourcesparseautoencoders,DBLP:journals/corr/abs-2410-19278,DBLP:journals/corr/abs-2409-14507} and proposed a comprehensive evaluation framework.

\wz{The interpretable features have been applied in various downstream tasks.
For instance, tracking activations of these features can be used to reveal internal model circuits~\cite{marks2024feature} and interpret model mechanisms~\cite{makelov2024sparse}.
Further, a number of studies manage to edit and control models by steering activations of SAE features.
}
% For downstream applications of features, researchers explored various possibilities. 
% Marks et al.\cite{marks2024feature} used SAE features to detect and analyze internal model circuits; Makelov et al.\cite{makelov2024sparse} utilized SAE to analyze indirect object recognition mechanisms; 
Wendler et al.\cite{wendler-etal-2024-llamas} studied the application of SAE features in multilingual translation. Templeton et al.\cite{templeton2024scaling} demonstrated how to use these features for model editing to intervene on specific concepts. Wu et al.\cite{wu2025interpretingsteeringllmsmutual} and Durmus et al.\cite{durmus2024steering} leverage features to steer language models for mitigating social biases and enhancing jailbreak defenses. Compared to prior work, ConceptViz bridges feature extraction and interpretation through organized feature mapping and borrowing the feature application technology of activation steering, enabling semantic validation of SAE features.\looseness=-1

\subsection{Visual Analytics for LLM Interpretability}

As machine learning models become increasingly complex, visual analytics has emerged as an essential tool for understanding and explaining these models \lhx{\cite{8371286, article, yang2024foundation}}. In the context of LLM interpretability, visual analytics approaches generally fall into two complementary paradigms: internal representation visualization and interaction-based external exploration. Internal representation visualization focuses on revealing the model's structure and computational processes. Early works such as CNNExplainer \cite{wangCNNExplainerLearning2020}, Seq2Seq-Vis \cite{8494828}, and LSTM-Vis \cite{8017583} visualized the internal mechanisms of different neural network architectures. With the rise of Transformer architectures, research shifted toward visualizing attention mechanisms, with tools like ExBERT \cite{tai-etal-2020-exbert} and BertViz \cite{DBLP:journals/corr/abs-1906-05714}, along with related works \cite{DeRose2020AttentionFA, yeh2023attentionviz, wangDodrioExploringTransformer2021}, displaying self-attention distributions. However, the academic community remains divided on the reliability of attention weights as a source of explanation \cite{clark-etal-2019-bert, jain-wallace-2019-attention, atanasova-etal-2020-diagnostic}. For overall LLM architecture, Gao et al. \cite{10297598} and Cho et al. \cite{cho2024transformer} designed interactive visualization tutorials for non-experts, while Tufanov et al.'s LM-TT toolkit \cite{ferrando2024information, tufanov2024lm} made the prediction process transparent, and Woodman et al. \cite{WOODMAN2024101238} visualized the LLM parameter space from a neuron-based perspective.

Interaction-based external exploration understands model behavior by manipulating inputs and analyzing outputs. PromptIDE \cite{9908590} supports visual experimentation with prompt engineering, while LMDiff \cite{strobelt-etal-2021-lmdiff} compares response differences between models given the same prompt. Recent research introduces more methods. Kahng et al.\cite{kahng2025comparator} enable interactive analysis of LLM side-by-side evaluation results, while Boggust et al.\cite{boggust2025compress} apply comparative assessment techniques to examine the impacts of model compression on LLM behavior. Coscia et al.~\cite{Coscia:2023:KnowledgeVIS} proposed methods for model explainability through fill-in-the-blank prompts and later extended this approach to LLM scoring explainability in educational contexts~\cite{Coscia_2024}. Wang et al.\cite{Wang_2023} explores model reasoning capabilities using external knowledge bases. These methods combine visualization with specific application scenarios, helping users build a comprehensive understanding of LLM capabilities and limitations. \lhx{While concept-based visualization approaches~\cite{ConceptExplainer, VisualConceptProgramming, DRAVA, Human-in-the-loop} have shown success in computer vision tasks, they have not been applied to LLM interpretability.} Unlike prior work, ConceptViz is the first to visualize SAE-decomposed discrete semantic features from LLM representations, enabling bidirectional concept-feature exploration while supporting validation on user-defined prompts, bridging both visualization paradigms for concept-to-feature analysis.

\section{Background and Tasks}
\subsection{Background on SAEs}
LLMs process text through transformer layers composed of attention mechanisms and feed-forward networks (MLPs). Attention heads capture contextual relationships between tokens, while MLPs apply non-linear transformations. The residual stream serves as the information pathway connecting these components throughout the network. Activations from attention heads, MLP layer, and residual stream all encode semantic information relevant to model interpretability.

SAEs~\lhx{\cite{shu2025surveysparseautoencodersinterpreting, bricken2023monosemanticity, cunningham2023sparseautoencodershighlyinterpretable}} function as an interpretability tool for these internal representations by decomposing dense neural activations into sparse, interpretable features. A typical SAE consists of an encoder ($W_{enc}$), a decoder ($W_{dec}$), and corresponding biases ($b_{enc}$ and $b_{dec}$). Given an input activation $x$, the SAE computes a hidden feature representation $z$ and reconstruction $\hat{x}$ through:

\begin{equation}
z = f(W_{enc}x + b_{enc}),
\end{equation}
\begin{equation}
\hat{x} = W_{dec}z + b_{dec},
\end{equation}

\noindent
where $f$ is a nonlinear activation function, typically ReLU or its variants\cite{Rajamanoharan2024JumpingAI}. SAEs attempt to reconstruct input activations by projecting them into a sparse, overcomplete feature basis using sparsity-inducing constraints. These constraints can be enforced either through $L_1$ regularization on the hidden activations or by directly limiting the number of non-zero features ($L_0$ sparsity).

The sparsity constraint is fundamental in encouraging the model to learn disentangled, mono-semantic features that potentially correspond to distinct concepts. While SAEs learn feature decompositions in an unsupervised manner, the interpretability and faithfulness of these learned features to the underlying LLM computation require systematic validation and analysis.

\subsection{Data Description}
\label{sec:data_description}

This section describes the foundational models and interpretive data resources used in ConceptViz, detailing both the technical infrastructure and methodological approaches for feature analysis.

\textbf{Model Information of LLMs and SAEs.}
\lhx{To support our study, we utilize \textit{GemmaScope}, an open-source suite of SAEs trained on the \textit{Gemma 2} series, including Gemma-2-2b, Gemma-2-9b, and Gemma-2-2b-it, Gemma-2-9b-it (instruction-tuned version). Specifically, we use the SAEs trained on the residual stream activations across Gemma-2-2b's 26 layers}, as researchers typically focus on residual stream features for more meaningful insights into model behavior \cite{templeton2024scaling}. The residual stream in Gemma-2-2b has an embedding dimension of 2304, which SAEs decompose into a 16384-dimensional sparse feature. This focused approach allows us to test our visualization system with the most relevant components for interpretability research, representing typical workflows in current LLM analysis.

%分三个 数据集 激活数据集 解释数据集 简单描述
% \textbf{Max Activation Text Segments and Automated Feature Explanations.}
% Researchers typically interpret SAE features by examining their max activation text segments across datasets. Our system uses the \textit{Neuronpedia} platform to provide high-activation examples from the \textit{Monology/Pile-Uncopyrighted} dataset and an automated explanation for every GemmaScope SAE feature. While these automated interpretations enhance accessibility, their reliability is limited by their dependence on only top-activating examples and potential inaccuracies in LLM pattern recognition, necessitating additional analytical approaches for robust feature understanding.

\textbf{Max Activation Text Segments and Automated Feature Explanations.}
\lhx{Our system utilizes pre-processed data from the \textit{Neuronpedia} platform~\cite{neuronpedia} derived from the \textit{Monology/Pile-Uncopyrighted} dataset, which contains 36,864 text segments of 128 tokens each.} For all GemmaScope SAE features, Neuronpedia calculates token-level activation values across these segments and provides representative text samples at different activation thresholds. The platform also supplies feature explanations generated by \textit{GPT-4o-mini}~\cite{bills2023language} based on patterns observed in top-activating text samples, offering preliminary insights into potential feature meanings and functions.

\subsection{Task Analysis}
%介绍workflow干啥，用户做什么行为

%下面T讲每个行为中会遇到什么问题，用户该怎么解决

% In the area of large language model interpretability, SAEs serve as powerful interpretability tools that decompose complex activation patterns within models into more interpretable feature representations. However, researchers exploring SAEs often face challenges in filtering, analyzing, and validating numerous trained models and features. Existing dashboard tools typically fail to provide a comprehensive, interactive exploration process with robust validation mechanisms, inadequately addressing the need for in-depth exploration of SAE internal representations. 
% \textcolor{gray}{(Introduce the research scope, target user) Our work aims to \dots .}
% To help users quickly identify features highly relevant to their concepts of interest and validate semantic consistency, ... 
% We present the analysis of tasks that support users in exploring concepts through SAEs, which will serve as the design goals of ConceptViz.
% Our work aims to support users who are interested in exploring concepts within LLMs through SAE techniques. 
\lhx{Our work aims to support LLM experts who are interested in exploring concepts within LLMs through SAE techniques.} Through a systematic task analysis based on an extensive literature review and in-depth exploration of existing interpretability tools~\cite{bloom2024saetrainingcodebase, neuronpedia}, 
we formulate the concept exploration task into a three-phase workflow.
In this workflow, users first formulate and refine queries to identify potentially relevant SAE models. They then navigate through feature spaces to discover and interpret meaningful features related to their concepts of interest. Finally, users validate their understanding by examining how these features activate across different inputs and testing how steering these features affects model outputs.
% we identified critical workflows and pain points in concept-driven SAE feature exploration. 
% To help users quickly identify features highly relevant to their concepts of interest and validate semantic consistency across different contexts, we designed a structured workflow with progressive levels of exploration and validation. 
We present the analysis of tasks that support users in exploring concepts throughout the workflow, which will serve as the design goals of ConceptViz.

\vspace{2pt}
\noindent
% 需求场景-》问题- 》解决方案 
% \textbf{R1. Recommend SAE Models Based on Concept Queries} % 精炼一些
\textbf{Identification of Valuable SAE.} At the start of concept exploration, researchers often need to carefully phrase the concept query and select a valuable SAE that well captures the concept\lhx{\cite{neuronpedia}}.
This process requires laborious trials for refinement of the concept query and cross validation between numerous candidate SAEs.

\begin{itemize}
% \item[\bf T1] \textbf{Query Optimization for Concept Search.}
% Concept queries need to be rephrased as they are vague or ambiguous.
% % Support users in refining initial concept queries that may be ambiguous or too broad. 
% For example, a query for ``apple'' which has dual meanings (technology company or fruit) requires more clarity for effective search. %与原来对比，指出原先方法的问题
% \textcolor{red}{(visualization requirement)}
% The system should provide intuitive visual representation on query quality and optimization suggestions.
% % , enabling users to compare original and optimized queries and understand how these refinements affect retrieval results.

\item[\bf T1] \textbf{Query Optimization for Concept Search.}
Concept queries need to be rephrased as they are vague or ambiguous\lhx{\cite{PromptMagician}}.
Users struggle with formulating effective queries for polysemous concepts, leading to either overly broad or excessively narrow search results. For instance, when exploring polysemous concepts like "interest" (which could refer to financial rates, personal hobbies, or attention), users must experiment with multiple phrasings to capture the intended meaning, a process that is time-consuming and lacks immediate feedback.
The system should provide intuitive visual feedback on query quality and suggest optimization alternatives, helping users understand how different formulations impact search results.

% \item[\bf T2] \textbf{Concept-based SAE Model Discovery.} 
% Candidate SAEs vary from their position in the LLM architecture.
% % Identify and rank relevant SAE models based on their training position and semantic similarity to the user's query. 
% SAEs from different model layers capture varying levels of abstraction—shallow layers often represent character-level features while deeper layers capture abstract concepts like sentiment. 
% \textcolor{red}{(visualization requirement)}
% The system should visualize similarity distributions between queries and SAE models across the network architecture, helping users locate the most promising models for exploration while reducing selection complexity.
% \end{itemize} 

\item[\bf T2] \textbf{Concept-based SAE Model Discovery.}
Candidate SAEs vary from their position in the LLM architecture.
Users lack efficient methods to identify which layers best represent their concepts of interest, forcing trial-and-error exploration across dozens of layer-specific SAE models without a coherent view of cross-layer concept evolution\lhx{\cite{balcells2024evolutionsaefeatureslayers, he2024llamascopeextractingmillions, lieberum-etal-2024-gemma}}. This results in missed insights and inefficient exploration processes.
The system should visualize similarity distributions between queries and SAE models across the entire network architecture, enabling users to identify the most relevant models and understand how concepts are represented at different abstraction levels.
\end{itemize}

\vspace{2pt}
\noindent
\textbf{Interpretation of SAE Features.} 
Upon selecting a SAE, researchers need to interpret its features related to the concept query. As the SAE features are numerous and cover various concepts, researchers typically initiate the process by filtering these features based on semantic similarity, subsequently conducting a thorough examination for detailed interpretation\lhx{\cite{templeton2024scaling, bricken2023monosemanticity}}.
% Some open-source tools\cite{neuronpedia} provide feature retrieval functionality, but present results in list format, requiring users to browse numerous features to locate those of interest. Even after selecting specific features, users must examine detailed information to build interpretation. 
% They need both macro-level overviews of concept-related features and well-organized micro-level details to quickly discover semantic patterns between and within features.

\begin{itemize}
\item[\bf T3] \textbf{Global Feature Visualization in Concept Space.}
SAE features lack intuitive spatial organization in concept space\lhx{\cite{bricken2023monosemanticity}}.
Researchers struggle with understanding how thousands of SAE features relate to each other and distribute across the human concept space. Without proper visualization support, users must mentally track connections between features, which becomes overwhelming as the feature count increases. The system should provide an intuitive spatial representation of features with highlighted regions relevant to user queries, enabling researchers to efficiently navigate the semantic landscape and identify feature clusters worthy of deeper investigation.

\item[\bf T4] \textbf{Detailed Feature Interpretation at Instance Level.}
Feature explanations require verification against actual activation patterns\lhx{\cite{bills2023language, paulo2025automatically}}.
When examining specific features, researchers face challenges in verifying the accuracy of feature explanations against actual behavior in text. The process of manually reviewing hundreds of activation examples is time-consuming and lacks structure, often leading to incomplete understanding of feature semantics. The system should provide organized presentation of activation examples, enabling researchers to discover nuanced feature behaviors and semantic patterns that might otherwise be missed.
\end{itemize}

\vspace{2pt}
\noindent
\textbf{Validation of Interpreted Concepts.}
After developing initial understandings of feature semantics, researchers need to empirically validate their hypotheses about SAE features\lhx{\cite{templeton2024scaling}}. 
\wz{As the inherent complexities of high-dimensional, noisy feature spaces, initial interpretations sometimes contain biases or overgeneralizations~\cite{elhage2022superposition}.}
This validation process is essential for confirming that identified features genuinely represent the concepts of interest and behave consistently across diverse contexts.

\begin{itemize}
\item[\bf T5] \textbf{Feature Activations Validation with Custom Inputs.}
Feature interpretations require testing against diverse real-world examples.
Researchers often struggle to verify whether identified features consistently respond to targeted concepts across diverse examples. Without systematic validation, users risk developing incorrect or overgeneralized interpretations based on limited samples. The system should display activation strengths for each token relative to target features and recommend alternative features with similar activation patterns when expected features fail to respond as anticipated.

\item[\bf T6] \textbf{Causal Relationships Verification with Output Steer.}
Correlation between features and concepts does not guarantee causation.
Establishing causal relationships between features and concepts presents a significant challenge, as correlation alone doesn't confirm that a feature truly represents a concept\lhx{\cite{shu2025inputactivationsidentifyinginfluential}}. Researchers need mechanisms to test whether manipulating specific features produces predictable changes in model outputs related to the concept of interest. The system should enable users to dynamically adjust input samples and feature activations while observing resulting changes in model behavior.
\end{itemize} 

\section{Methods}

\begin{figure*}[t]
\centering
\includegraphics[width=\textwidth]{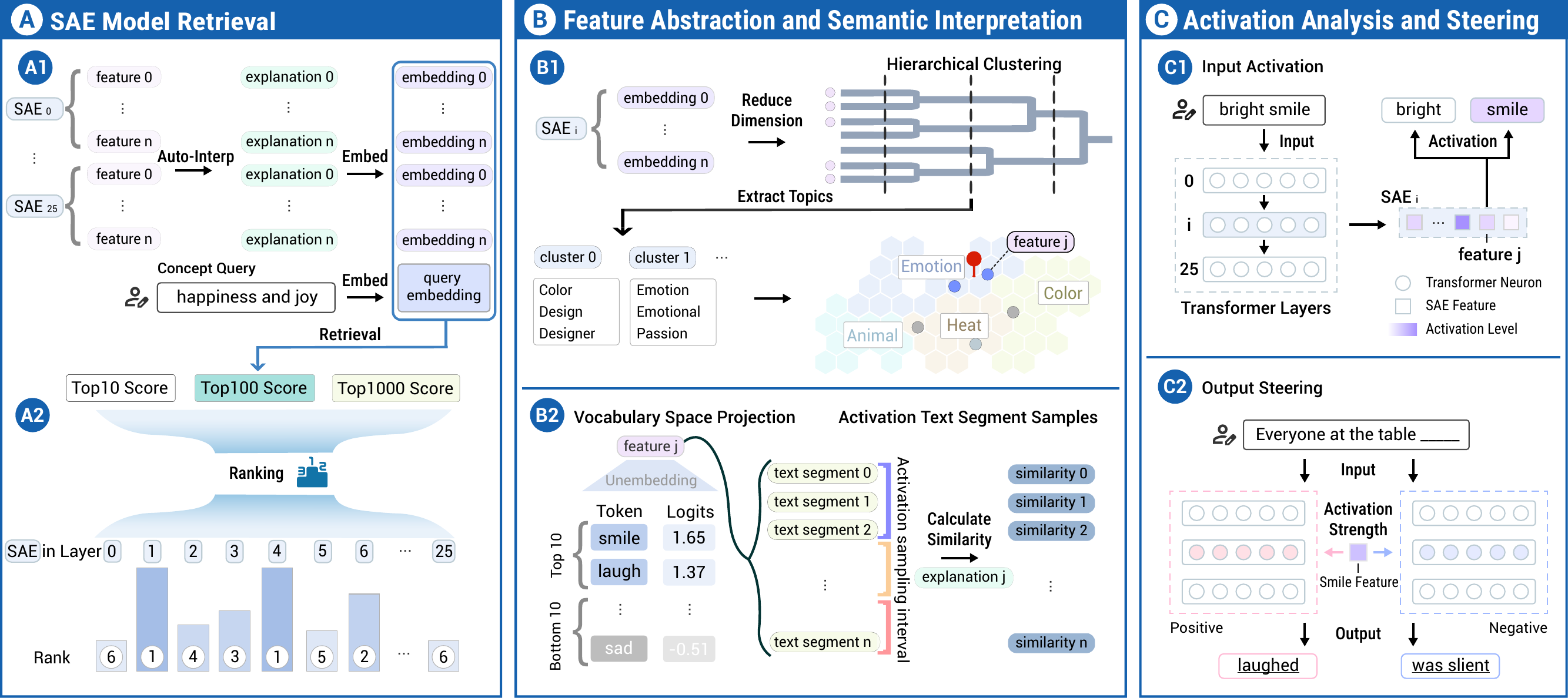}
\caption{Our system's three technical components: (A) Concept-based SAE Model Retrieval identifies relevant models by matching user queries with feature explanations, computing multi-threshold similarity rankings, and prioritizing consistently high-performing SAEs. (B) Feature Abstraction and Semantic Interpretation provides both global context through hierarchical clustering and topic visualization (B1), and detailed feature understanding through vocabulary projection and activation pattern analysis (B2). (C) Activation Analysis and Steering enables empirical validation through custom input testing and causal verification by manipulating feature activations to observe resulting output changes.}
\label{fig:method}
\end{figure*}

As illustrated in~\cref{fig:method}, our system consists of three components: (A) a concept-based SAE model retrieval mechanism that identifies relevant representations, (B) a multi-level feature abstraction and semantic interpretation framework that organizes and explains features, and (C) an interactive activation analysis and steering approach that enables empirical validation of feature behaviors.

\subsection{SAE Model Retrieval}
Our model retrieval system enables efficient concept-driven selection of SAE models (\cref{fig:method}A). We calculate semantic similarity between user queries and SAE features using OpenAI's \textit{text-embedding-3-large} model, which encodes both feature explanations and queries into 3072-dimensional vectors (\cref{fig:method}A$_1$). The system identifies the most concept-relevant SAE models through a multi-level ranking approach (\cref{fig:method}A$_2$). For each query, we compute cosine similarity scores between the query embedding and all feature embeddings across all SAE models. We then identify which SAE models contribute the most features to the Top-K most similar features, where $\mathcal{K} = \{10, 100, 1000\}$. This multi-threshold approach ensures we capture both highly specific matches (K=10) and broader semantic relationships (K=1000). Our ranking algorithm computes a balanced recommendation score by averaging each SAE model's rank position across these different thresholds:

\begin{equation}
\text{AvgRank}(SAE_i) = \frac{1}{|\mathcal{K}|} \sum_{K \in \mathcal{K}} \text{Rank}(SAE_i, \text{Top-K}).
\end{equation}

\vspace{2pt}
\noindent
This approach prioritizes SAE models that consistently rank well across different thresholds, mitigating the influence of any single metric and providing robust recommendations that balance precision and recall in matching user concepts to SAE feature spaces.

\subsection{Feature Abstraction and Semantic Interpretation}

To facilitate exploration of thousands of SAE features, we developed two complementary approaches (\cref{fig:method}B): a hierarchical semantic clustering method for organizing features into concept-related groups (\cref{fig:method}B$_1$), and a detailed feature analysis method (\cref{fig:method}B$_2$) for interpreting individual feature semantics. The clustering approach helps users navigate the feature landscape efficiently, while the semantic interpretation techniques provide deeper insights into feature behaviors and meanings. Together, these methods support comprehensive feature interpretation from both macro and micro perspectives.

\vspace{2pt}
\noindent
\textbf{Dimensionality Reduction.} We apply UMAP to reduce the 3072-dimensional feature embeddings to a 2D space, using cosine distance as the metric ($n_{neighbors}=15$, $min_{dist}=0.3$). UMAP preserves both local relationships and global topological structure while providing computational efficiency.

\vspace{2pt}
\noindent
\textbf{Hierarchical Clustering.} 
For each SAE model, we employ Agglomerative Hierarchical Clustering at multiple granularity levels (10, 30, and 90 clusters) to construct a semantic organization of features. We select Ward linkage as our clustering criterion, which minimizes within-cluster variance:

\begin{equation}
d(u,v) = \sqrt{\frac{|v|+|s|}{T} d(v,s)^2 + \frac{|v|+|t|}{T} d(v,t)^2 - \frac{|v|}{T} d(s,t)^2},
\end{equation}

% \vspace{2pt}
\noindent
where $u$ is a new cluster formed by merging clusters $s$ and $t$, $v$ is any other cluster, $T = |v|+|s|+|t|$, and $|\cdot|$ denotes the number of samples in a cluster. Ward's method creates balanced clusters of similar sizes, avoiding the chaining effect problem of single-linkage clustering.

\vspace{2pt}
\noindent
\textbf{Topic Extraction.} To enhance interpretability, we leveraged a class-based TF-IDF approach to extract representative keywords for each cluster. 
Our method treats all feature explanations within each cluster as a composite document:

\begin{equation}
c\text{-}TF\text{-}IDF_{t,c} = TF_{t,c} \times IDF_t = \frac{f_{t,c}}{\sum_{t' \in c} f_{t',c}} \times \log\frac{|C|}{|\{c \in C: t \in c\}|},
\end{equation}

% \vspace{2pt}
\noindent
where $f_{t,c}$ is the frequency of term $t$ in cluster $c$, and $|C|$ is the total number of clusters. We extract the top 5 terms with highest c-TF-IDF scores as conceptual labels for each cluster after filtering domain-specific stop words.

\vspace{2pt}
\noindent
\textbf{Semantic Interpretation.} Our feature semantic interpretation addresses limitations in current automatic explanation methods by providing vocabulary space projections and text segments activation patterns (\cref{fig:method}B$_2$). We extract feature projections in vocabulary space by computing $W_u \cdot W_{dec}[feature]$ (where $W_u$ is the unembedding matrix), revealing token prediction preferences during feature activation. Leveraging pre-computed activation data with stratified sampling, we analyze high-activation instances and samples across various activation intensities. Inspired by EleutherAI's explanation scoring method\cite{paulo2025automatically}, we compute cosine similarities between text embeddings and feature explanation embeddings to quantify semantic alignment. This similarity metric, when examined alongside activation values, helps identify potential explanation-behavior discrepancies, particularly instances exhibiting mismatches between activation values and semantic similarity.

\subsection{Activation Analysis and Steering}

\vspace{2pt}
\noindent
\textbf{Custom Input Activation Analysis.} 
Our verification method analyzes how features respond to user-defined text (\cref{fig:method}C$_1$). Following the same approach used to compute feature activations on pre-collected text segments, we implemented this method for user-defined text. When users input text, the system performs a forward pass through the model to obtain intermediate activations, maps them to the feature space via the SAE encoder, and records the activation strength of target features at each token position. This process reveals how features respond to specific linguistic contexts in real-time. Beyond single feature analysis, we provide a co-activation feature retrieval function that calculates activation values of all features for user-selected tokens and returns other features that similarly exhibit high activation for the same tokens. This capability helps users discover functionally related feature sets that may respond to similar semantic or syntactic patterns.

\vspace{2pt}
\noindent
\textbf{Causal Verification through Activation Steering.} To establish causal relationships between features and concepts, we implement Activation Steering techniques that precisely control feature activation (\cref{fig:method}C$_2$). We extract the steering vector for a feature from SAE ($W_{dec}[feature]$) and selectively adjust the activations of the target layer during model inference:

\begin{equation}
activations += steering\_strength \times steering\_vector.
\end{equation}

\vspace{2pt}
\noindent
This method allows users to observe the causal impact of feature activation changes on model output by setting different steering strengths (positive values enhance feature influence, negative values suppress it). By comparing outputs with and without steering, users can verify the association between features and hypothesized concepts through direct causal intervention, providing empirical evidence of feature semantics.

\section{System Design}

We have developed a visual analytics system that leverages Sparse Autoencoder (SAE) models and their feature explanations to help researchers explore the SAE feature space based on concepts and validate concept-feature matches. In this section, we introduce the system interface and detail the key components supporting the analytical workflow.

\subsection{User Interface}

The system interface (\cref{fig:teaser}) consists of six coordinated views that form a complete analytical workflow. The Concept Query View and SAE Discovery View help users locate SAE models with features related to target concepts. The Feature Explorer View and Feature Details View enable exploration of feature distributions and detailed semantic information. The Input Activation View and Output Steering View support the validation of connections between features and concepts.

\vspace{2pt}
\noindent
\textbf{Concept Query View (\cref{fig:teaser}A)} serves as the entry point for feature exploration (\textbf{T1}). Users describe concepts of interest using natural language, and the system calculates semantic similarity between all feature explanations and the query. A similarity distribution chart of the top 2,000 features provides an initial overview of the match between the concept query and feature semantics. This view supports query refinement through suggestions by \textit{GPT-4o}, helping users address ambiguity and optimize search effectiveness.

\vspace{2pt}
\noindent
\textbf{SAE Discovery View (\cref{fig:teaser}B)} enables users to select appropriate SAE models based on concept-relevance metrics (\textbf{T2}). The overview bar above in this view displays the distribution of concept-relevant features across different network layers, presenting Top 10, Top 100, and Top 1000 scores for each layer in the list below. This layer-based organization helps users identify SAEs that balance model architectural considerations with concept relevance, allowing them to select the most suitable SAE for their specific analysis needs. Users can directly select the SAE of interest from this view to proceed with feature exploration.

\vspace{2pt}
\noindent
\textbf{Feature Explorer View (\cref{fig:teaser}C)} presents a global visualization of SAE features (\textbf{T3}). Through embedding and dimensionality reduction, all feature explanation vectors are mapped to a two-dimensional semantic space. The system highlights query-relevant features and represents the query's position with a red pin. Hierarchical clustering results are displayed as a color-coded map with representative topic terms at cluster centers. We employ HexBin techniques for cluster semantics and level transitions through zooming, offering multi-granularity semantic views. \lhx{HexBin aggregation reduces visual clutter from thousands of feature points while maintaining interactive performance.}

\vspace{2pt}
\noindent
\textbf{Feature Details View (\cref{fig:teaser}D)} presents comprehensive information about selected features (\textbf{T4}). The left panel (D1) shows the feature's projection in vocabulary space, mapping words with the highest and lowest probabilities. The right panel (D2) displays an activation-similarity matrix of sampled text segments, highlighting cases where activation patterns differ from explanation predictions. Adjacent to this matrix, a bar chart summarizes the maximum activation tokens for each text segment, supporting interactive selection. When users select regions in the matrix, the bar chart updates to show token distributions from the selected segments. Users can also view complete sentences for context through the View Details option. This multi-faceted approach reveals discrepancies between explanations and actual feature behavior.

\vspace{2pt}
\noindent
\textbf{Input Activation View (\cref{fig:teaser}E)} allows users to verify feature responses to custom inputs (\textbf{T5}). Users can enter text samples, and the system displays the activation distribution across tokens for the target feature. For selected tokens with high activation, the system recommends other features with similar activation patterns by highlighting bubble sets displayed in the Feature Explorer View. This enables users to discover functionally related feature sets that respond to similar semantic or syntactic patterns. The view supports instance-level validation of whether features activate as expected for specific concepts.

\vspace{2pt}
\noindent
\textbf{Output Steering View (\cref{fig:teaser}F)} enables causal verification of feature-concept relationships (\textbf{T6}). The Activation Steer module allows users to input prompts and generate multiple output variants with different feature activation strengths simultaneously. By comparing outputs across different steering branches, users can observe how manipulating feature activations affects model outputs. This view facilitates empirical verification of whether steering features toward or away from specific concepts produces corresponding changes in model behavior, establishing causal relationships between features and concepts.

% These six views form an integrated analytical framework, supporting a complete workflow from concept query refinement to SAE selection, feature exploration, detailed feature analysis, and concept validation through both input-based and output-based verification methods.

\subsection{Hierarchical Feature Visualization}

In the Feature Explorer View, we implement a hierarchical color coordination scheme to support seamless exploration across different granularity levels and address the color consistency challenge in multi-level cluster visualization \wz{(see \cref{fig:teaser}C)}. When users switch between clustering levels through zooming, color assignment continuity is crucial for maintaining cognitive coherence. Assigning colors to clusters independently at different levels results in excessive color differences between parent and child clusters that increase cognitive load. Inspired by Liu et al.\cite{10669934}, we implement a hierarchical color coordination mechanism operating in HSL color space. This approach assigns a base hue to each top-level cluster, with child clusters inheriting and controllably varying their parent's hue. Specifically, child cluster hues are determined by:

\begin{equation}
h_{child} = (h_{parent} + \Delta h) \bmod 1.0,
\end{equation}

\vspace{2pt}
\noindent
where $\Delta h$ is a controlled random variation ensuring child cluster hues remain close to their parent's. Simultaneously, we enhance discriminability between child clusters by adjusting lightness, allowing child clusters to maintain similar hues to their parents while exhibiting sufficient visual distinction.

To prevent excessive similarity between same-level clusters, we introduce a minimum color difference constraint, ensuring the Euclidean distance between any two same-level clusters in HSL space exceeds a preset threshold:

\begin{equation}
d(c_i, c_j) > \tau, \forall i \neq j,
\end{equation}

\vspace{2pt}
\noindent
where $\tau$ is the minimum color difference threshold, and $c_i$ and $c_j$ represent HSL color triplets.

\subsection{Anomalous Activation Text Segments Visualization}
To help users understand feature behavior beyond automatic explanations, we designed an activation-similarity matrix that reveals potential inconsistencies between feature activations and explanations. In the Feature Details View, we compare each sampled sentence's maximum activation value with the cosine similarity between the sentence's text embedding and the feature explanation embedding \wz{(see \cref{fig:teaser}D)}. Ideally, activation strength should correlate positively with semantic similarity--high-activation sentences should exhibit high semantic relevance to the explanation.

\begin{figure}[htbp]
\centering
\includegraphics[width=1.0\columnwidth]{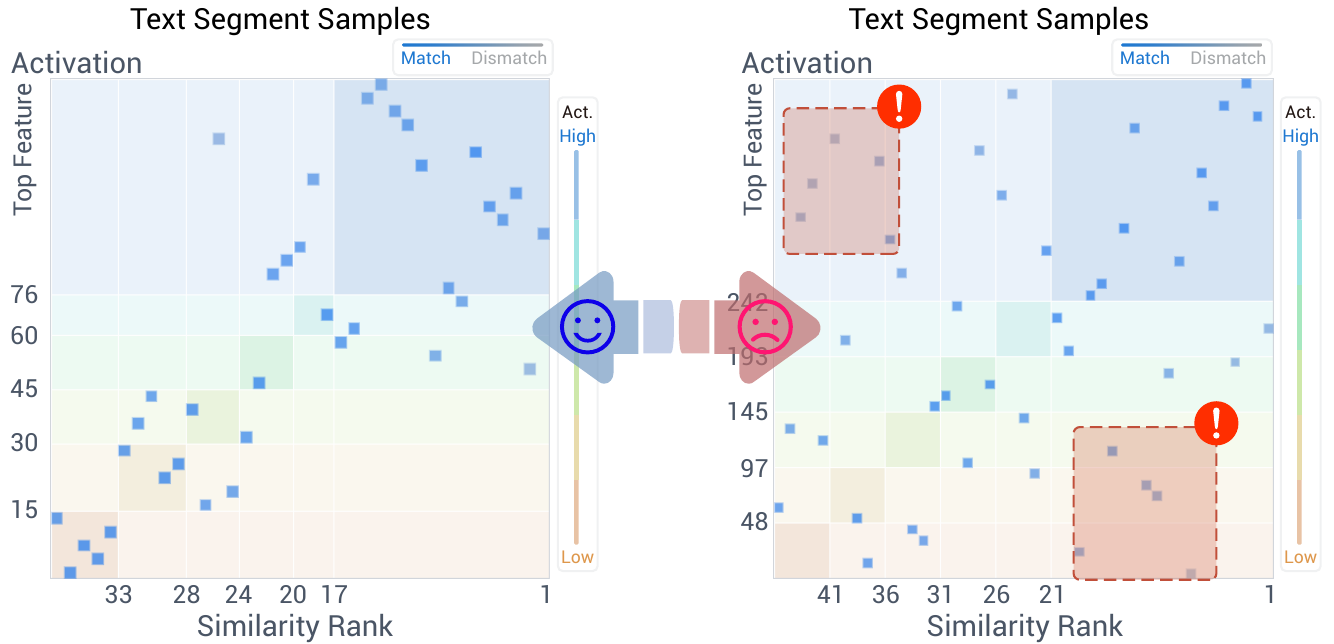}
\caption{Activation-similarity matrix visualization: Left: Well-explained feature with correlation between activation strength and semantic similarity (samples align along diagonal). Right: Poorly explained feature showing discrepancies - high-activation samples with low similarity (upper left) and high-similarity samples with low activation (lower right), revealing explanation limitations.}
\label{fig:uglychart}
\end{figure}

\lhx{This matrix is visualized as a two-dimensional plot as shown in \cref{fig:uglychart}.  The x-axis represents the ranking of sentences by semantic similarity to the feature explanation, while the y-axis represents their ranking by maximum activation value. Each cell represents a sampled text segment, with its position revealing the degree of match between activation strength and semantic relevance rankings. The matrix uses a background color gradient from blue (high activation) to red (low activation) to encode activation value regions, as indicated by the color legend. Text segments falling along the color-emphasized diagonal regions, where high activation corresponds to high semantic similarity, indicate cases where activation and semantic similarity rankings approximately align, as shown in the left figure. This alignment suggests the feature explanation reasonably captures activation behavior. Deviations from the diagonal represent potential anomalies. Segments in the upper left region (blue background, high activation) with low semantic similarity potentially indicate feature behaviors not captured by the automatic explanation. Conversely, segments in the lower right show high semantic relevance but low activation (red background), suggesting the explanation covers concepts that don't consistently trigger the feature.}

\begin{figure*}[t]
\centering
\includegraphics[width=\textwidth]{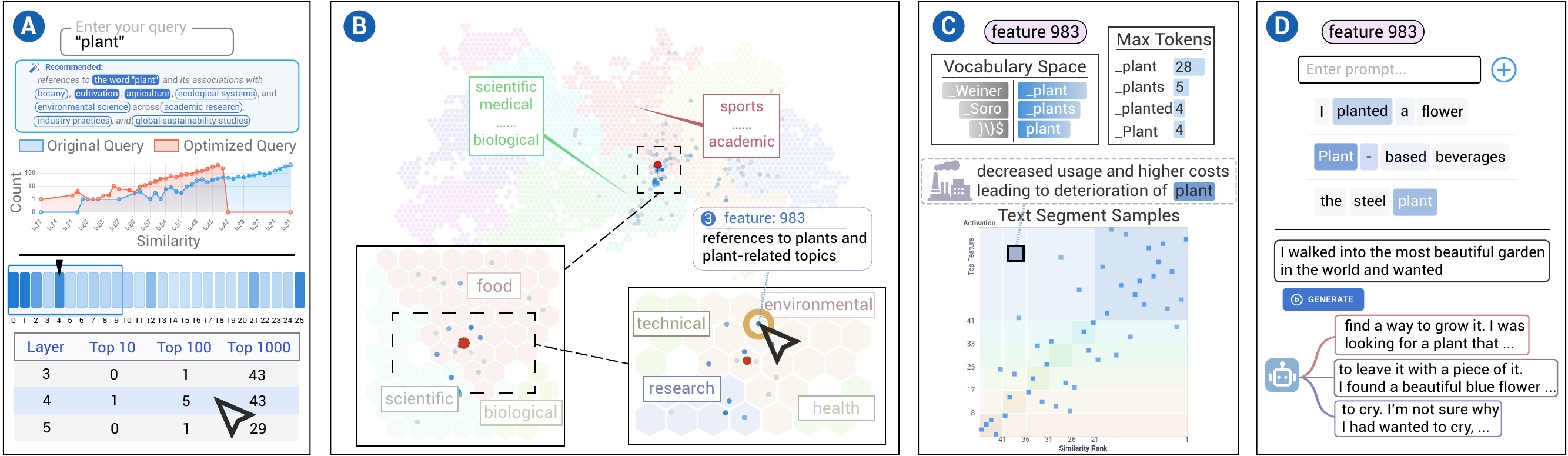}
\vspace{-4mm}
\caption{\lhx{A user exploring plant-related features: (A) The user queries ``\textit{plant}'', receives an optimized suggestion, then examines SAE relevance rankings with Layer 4 showing higher relevance. (B) While browsing feature clusters, the user identifies a relevant feature near the \textcolor{olive}{environmental} region. (C) After selecting feature 983, the user examines its vocabulary space and activation matrix, revealing plant-related associations. (D) The user analyzes activations and validates through steering, observing effects on garden text generation.}}
\label{fig:scenario_1}
\vspace{-4mm}
\end{figure*}

\section{Usage Scenarios}
To demonstrate our system's analytical capabilities, we present two usage scenarios \lhx{completed by two domain experts using the datasets described in Sec \ref{sec:data_description}.} First, the user explores plant-related features, showcasing how our system helps understand concept and feature semantic associations. \lhx{Second, the user investigates the superhero concept, highlighting our system's ability to reveal meaningful token-feature associations while distinguishing semantically relevant features from coincidental activations.}

\subsection{Exploring and Validating Plant-Related Features}

In this scenario, the user begins by entering ``\textit{plant}'' in the Concept Query View (\cref{fig:scenario_1}A). The system calculates semantic similarity between feature explanations and both original and optimized queries, displaying the distribution graph. Observing higher similarity counts with the optimized query, the user refines their search to ``\textit{words related to plant and its associations with cultivation, agriculture}''. After finalizing the query, the user examines the SAE Model Retrieval View (\cref{fig:scenario_1}A), which displays a heat map of semantic similarity across all SAE layers. The visualization reveals stronger semantic matches (darker blue bars) concentrated in the earlier layers. Based on this insight, the user selects the fourth layer SAE for detailed exploration.

%In the Feature Explorer View, the user observes query-relevant features (blue points) distributed across four cluster regions, with higher density near the "environmental" keyword cluster. A particularly dark blue point catches attention, with its feature description showing "references to plants and plant-related topics." The user selects this point for detailed analysis. The Feature Details View displays vocabulary space projections and maximum token activation statistics on the left, both confirming the feature's association with lexical variations such as "plant", "plants", "planted", and "Plant". The activation sample panel further validates this semantic alignment. However, in the activation-similarity matrix, a gray point indicating high activation but low semantic similarity suggests potential limitations in the automated feature explanation. Upon inspection, the user discovers that this sample contains the word "plant" referring to a factory rather than vegetation, indicating that the feature primarily recognizes the lexical token "plant" while associating it with botanical meanings.

In the Feature Explorer View, the user observes query-relevant features (blue points) distributed across several semantic regions in the concept space (\cref{fig:scenario_1}B). Starting at a high-level view, user identify broad conceptual areas like \textcolor{olive}{scientific} and \textcolor{olive}{sports}. Drilling down into the \textcolor{olive}{scientific} region reveals intermediate clusters focused on \textcolor{olive}{food} concepts. Further exploration uncovers finer-grained semantic groups including \textcolor{olive}{environmental} and \textcolor{olive}{research} topics. Within the \textcolor{olive}{environmental} cluster, the user discovers a highlighted feature with the description \textit{"references to plants and plant-related topics"}. The user selects this point for detailed analysis(\cref{fig:scenario_1}C). The Feature Details View displays vocabulary space projections and maximum token activation statistics on the left, both confirming the feature's association with lexical variations such as ``\textit{plant}'', ``\textit{plants}'', ``\textit{planted}''. The overview of activation-similarity matrix further validates this semantic alignment. However, in the activation-similarity matrix, a gray point indicating high activation but low semantic similarity suggests potential limitations in the automated feature explanation. Upon inspection, the user discovers that this sample contains the word ``\textit{plant}'' referring to a factory rather than vegetation, indicating that the feature primarily recognizes the lexical token \ecl{plant} while associating it with botanical meanings.

To validate this hypothesis(\cref{fig:scenario_1}D), the user proceeds to the Custom Input Activation View, inputting text containing token \ecl{plant}. The activation pattern analysis confirms the earlier findings. The user then employs the Activation Steering View with the prompt ``\textit{I walked into the most beautiful garden in the world and wanted}'' and observes completion results across different feature activation levels. The experiments reveal that increasing feature activation produces outputs focused on plants and botanical vocabulary; neutral activation maintains garden-related content with fewer specific plant references; while suppressing activation shifts away from plant-related content toward environmental and emotional experiences. Through this analysis process, the user validates that the feature indeed correlates with both the word ``\textit{plant}'' and its botanical meaning, while discovering nuanced activation 
patterns beyond the scope of the automated explanation.

\lhx{\subsection{Exploring Superhero-Related Features}}

\begin{figure*}[t]
\centering
\includegraphics[width=\textwidth]{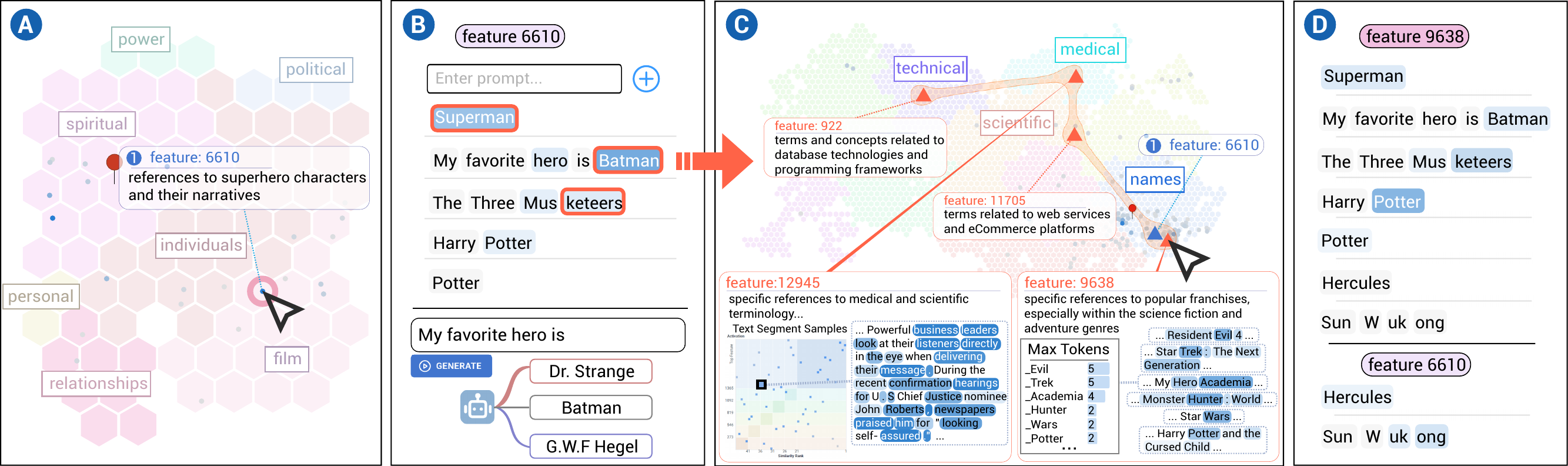}
\caption{\lhx{A user investigating superhero-related features: (A) The user identifies feature 6610 within the \textcolor{olive}{individuals} cluster, referencing superhero characters. (B) Input activation shows strong responses to superhero names, while steering demonstrates control over character generation. (C) Most related features are distant with poor labels, but feature 9638's proximity captures user interest. (D) Validation reveals feature 9638 activates for entertainment franchises like \ecl{Harry Potter} but not traditional heroes like \ecl{Hercules}.}}
\label{fig:scenario_2}
\vspace{-4mm}
\end{figure*}

\lhx{In this scenario, the user begins by querying ``\textit{superhero}'' in the Concept Query View (\cref{fig:teaser}A) and selects an appropriate SAE model from layer 11 (\cref{fig:teaser}B). The Feature Explorer View (\cref{fig:scenario_2}A) displays query-relevant features distributed across clusters labeled as \textcolor{olive}{power}, \textcolor{olive}{spiritual}, and \textcolor{olive}{individuals}. Feature 6610, described as ``\textit{references to superhero characters and their narratives}'', captures the user's attention due to its position within the \textcolor{olive}{individuals} cluster.}

\lhx{In the Input Activation View (\cref{fig:scenario_2}B), the user tests this feature with various superhero-related prompts. The visualization reveals strong activation for superhero names, demonstrating the feature's ability to recognize popular superhero characters. For validation, the user employs the Output Steering View with the prompt ``\textit{My favorite hero is}'' and observes how different activation strengths influence completions: high activation generates superhero names like ``\textit{Dr. Strange}''  while lower activation produces more general character references.}

\lhx{The user then selects superhero tokens (e.g., \ecl{Superman}, \ecl{Batman}) to identify other features with similar activation patterns (Fig. 6C).
% The user then explores (\cref{fig:scenario_2}C) whether other features exhibit similar behavior by selecting superhero tokens like \ecl{Superman}, \ecl{Batman}, and \ecl{keteers} to identify additional features responding to these tokens. 
While three other features show activation for these superhero tokens, their semantic cluster labels (\textcolor{olive}{technical}, \textcolor{olive}{medical}, and \textcolor{olive}{scientific}) immediately suggest they may not represent superhero concepts coherently. Upon examining feature 12945, the user confirms it exhibits poor activation-to-semantic coherence, with text segments triggering highest activation spanning diverse, unrelated concepts, indicating this feature lacks clear semantic specificity. However, feature 9638's proximity to feature 6610 in the feature space captures the user's attention. Through maximum token activation statistics, the user observes this feature relates to heroic movies, demonstrating its connection to superhero concepts.}

\lhx{For more detailed validation (\cref{fig:scenario_2}D), the user discovers that feature 9638 activates for entertainment franchises like \ecl{Harry Potter} but shows no response to traditional heroes from mythology and folklore like \ecl{Hercules} and \ecl{Sun Wukong}. 
Based on this key finding, the user concludes that feature 6610 captures a general heroic concept, while feature 9638 is specific to characters from entertainment franchises.}

\vspace{4pt}
\section{User Study}
We conducted a formal user study to evaluate ConceptViz. 
Specifically, we aimed to evaluate (1) the effectiveness of \wz{each system component (\cref{fig:teaser}A-F)} in helping users understand and explore features, and (2) the overall usability and workflow design of the system.

\subsection{Participants}
We recruited 12 participants (P1-P12) from a local university through campus forums and online channels. All participants were undergraduate and master's students in computer science who had previously participated in LLM research. Three of them had experience in AI explainability research. All participants had some understanding of SAEs, with most having a solid grasp of SAE fundamental knowledge. Each participant received \$20 as compensation for their time.

\subsection{Procedure and Tasks}

\vspace{2pt}
\noindent
\textbf{Training Phase (20 minutes):} We first introduced the research background and motivation to the participants. We then collected demographic information from participants and obtained their consent to record their system usage, results, and responses for further analysis. Following this, we demonstrated the system operation and gave participants sufficient time to familiarize themselves with the various functions and interaction methods of the system.

\vspace{2pt}
\noindent
\textbf{Targeted Task (20 minutes):} To ensure all participants could evaluate the system on an equal basis, we asked them to complete a standardized guided task: using the system to identify the one feature most relevant to ``negative emotions''. This task was designed to test the basic functionality of the system and the participants' proficiency with it.

\vspace{2pt}
\noindent
\textbf{Open-ended Exploration (20 minutes):} In this phase, we encouraged participants to choose a concept domain of interest (e.g., food, history, or media) and freely explore using the ConceptViz system. The participants' goal was to discover and validate at least one meaningful feature related to their chosen concept. This task was designed to evaluate the system's capability to support open-ended exploration and discovery.

\vspace{2pt}
\noindent
\textbf{Questionnaire and Interview (20 minutes):} After the experiment, participants completed questionnaires based on a 5-point Likert scale, evaluating the system's effectiveness, usability, and overall satisfaction (\cref{fig:result}). 
Subsequently, we conducted semi-structured interviews to collect detailed feedback from participants regarding the effectiveness of each system component, the overall workflow, and the system usability.

\subsection{Results Analysis}
\begin{figure}[t]
  \centering
  \includegraphics[width=1.0\columnwidth]{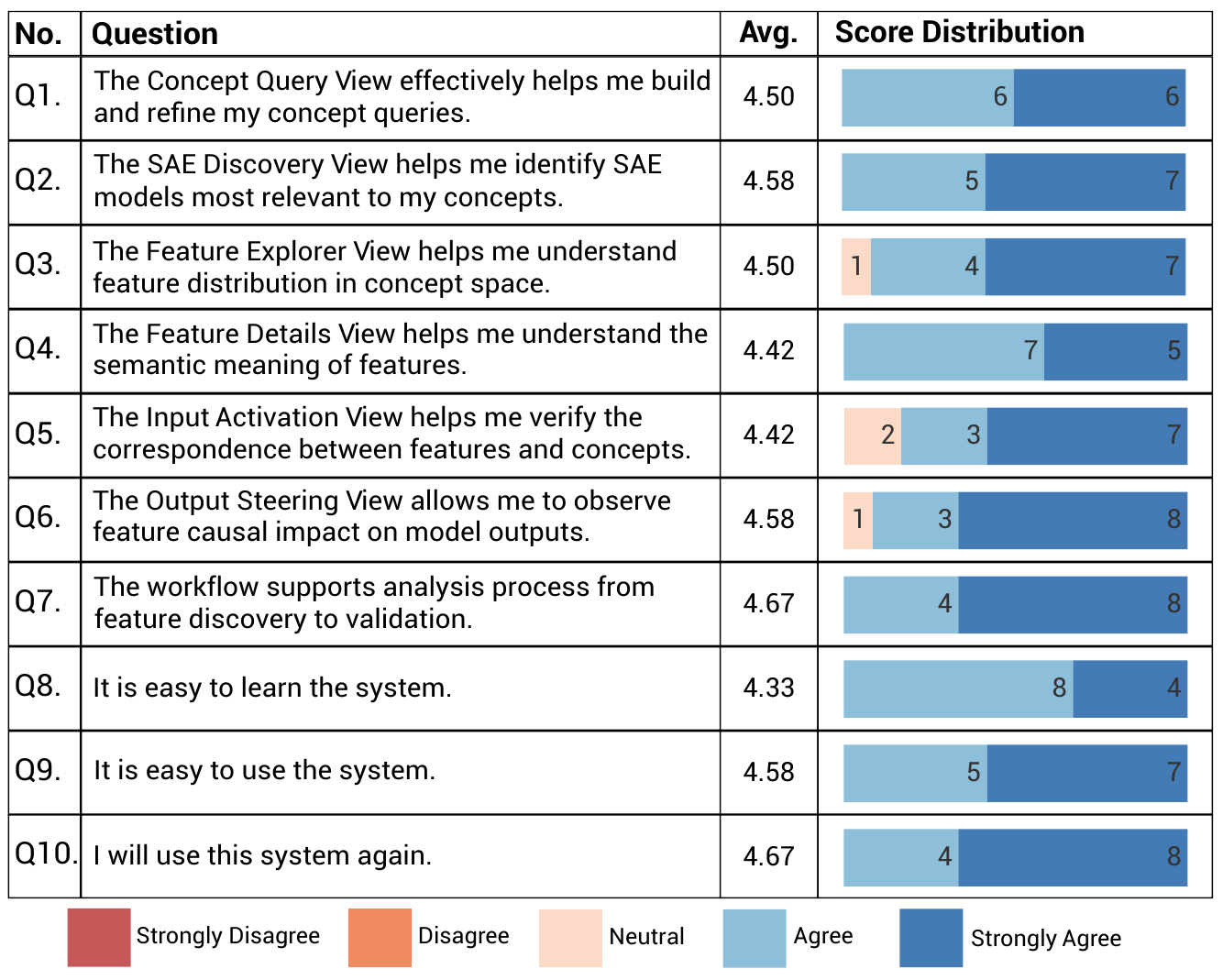}
  \caption{The results of the questionnaire regarding the effectiveness and
usability of the visual system and workflow.}
  \label{fig:result}
  \vspace{-4mm}
\end{figure}

% All participants completed the training phase, guided task, and open-ended exploration, experiencing all system functionalities. 
\wz{Based on the collected feedback and questionnaire ratings (\cref{fig:result}), we analyzed the effectiveness of each system component (\cref{sec:result_1}),
and summarized the overall usability along with notable user behavior patterns and potential areas for improvement (\cref{sec:result_2}).}

\subsubsection{Effectiveness of System Components}
\label{sec:result_1}
\wz{Overall, all of our system components received ratings higher than 4 on average (Q1--Q6), showing users' satisfaction on their effectiveness.}

\vspace{4pt}
\noindent
\wz{\textbf{Identification.}
Participants found the Concept Query View (\cref{fig:teaser}A) helpful for building and refining concept queries (Q1, $\mu=4.50$).}
P6 highlighted that ``\textit{it provides valuable query improvement suggestions when concepts are ambiguous}.'' 
\wz{For the SAE Discovery View (\cref{fig:teaser}B), participants generally found the suggestions effective (Q2, $\mu=4.58$),} with P3 noting that ``\textit{it allows me to quickly identify the SAE layer most relevant to my query concept},'' significantly reducing the time needed to find appropriate features. In addition, P9 mentioned that effectively using this functionality requires understanding how different SAE layers capture semantics, noting that ``\textit{lower-level SAEs detect concrete patterns while higher-level ones capture abstract concepts}.''

\vspace{4pt}
\noindent
\wz{
\textbf{Interpretation.}
The Feature Explorer View (\cref{fig:teaser}C) received high ratings for helping users understand the semantic distribution of features within SAEs (Q3, $\mu=4.50$).}
P3 indicated it clearly partitions different concept areas through semantic clustering, making it easier to identify feature groups related to specific topic, while P10 found the topic modeling functionality intuitively displays feature semantic categories through meaningful spatial organization. Some participants suggested improvements to the visualization aspects, with P7 noting ``\textit{color transitions between feature clusters could be smoother to better indicate semantic relationships},'' and P4 observing that in certain cases, query-relevant and irrelevant features (blue and gray points) aren't sufficiently separated in the spatial layout', suggesting that ``\textit{the dimensionality reduction might benefit from incorporating query information}.''

\wz{Participants acknowledged the Feature Details View (\cref{fig:teaser}D) helpful for understanding feature semantics (Q4, $\mu=4.42$).}
P10 valued observing text segment samples where activation values significantly diverge from similarity, which reveals discrepancies between feature activation patterns and feature explanations, while P8 appreciated how the combination of vocabulary space visualization and max activation token statistics enables rapid comprehension of feature semantics. P6 suggested future versions could benefit from more automatically recommending feature explanations and representative sentences to accelerate the feature understanding process, especially for complex cases.

\vspace{4pt}
\noindent
\wz{
\textbf{Validation.}
The Input Activation View (\cref{fig:teaser}E) and Output Steering View  (\cref{fig:teaser}F) received positive feedback for their functionalities. Input Activation View was rated highly for verifying feature-concept relationships (Q5, $\mu=4.42$)}, with P9 praising its highly interactive interface design that allows direct testing of features against different input samples and P3 noting the benefits of combining token-associated features bubble sets with the Feature Explorer View to efficiently validate semantic alignment. 
\wz{Output Validation was appreciated for observing feature impacts on model generation (Q6, $\mu=4.58$),} with P1 describing it as ``\textit{the system's most excellently designed component for revealing how features influence text generation outcomes},'' and P6 indicating it helps verify feature function at the model level by showing how activated features alter the LLM output. P4 suggested that the Activation Steer component was promising with further optimization, as users sometimes need multiple attempts to discover meaningful insights.

\subsubsection{Overall Workflow and System Usability}
\label{sec:result_2}
Participants were overall satisfied with workflow and system usability.

\vspace{2pt}
\noindent
\wz{
\textbf{Workflow.}
Participants considered the three-stage workflow design reasonable (Q7, $\mu=4.67$),} supporting complete analysis from SAE identification to validation. P4 praised its ``\textit{macro-to-micro analysis path},'' while P9 noted the workflow has ``\textit{clear logical transitions between stages}.'' P6 evaluated that ``\textit{the views form a tightly complementary analysis loop rather than merely sequential relationships}.''

\vspace{4pt}
\noindent
\textbf{System Usability.}
\wz{Regarding learnability (Q8), most participants found the system easy to learn ($\mu=4.33$),} despite requiring some prerequisite knowledge. P4 considered the interface intuitive and easily mastered, while P8 suggested providing an overview visualization of Transformer model architecture to help users connect interface elements with underlying concepts. 
\wz{In terms of usability (Q9), participants found the system easy to use ($\mu=4.58$).} 
Most participants mentioned the exploration process was engaging and enjoyable. P8 noted that the system could be valuable beyond professional researchers, extending to language model enthusiasts who want to understand how these models work.
\wz{Overall, all participants indicated a willingness to use the system again (Q10, $\mu=4.67$),} suggesting significant value in supporting the exploration of internal features and learning meaningful concepts in LLMs.

\section{Discussion}
% \subsection{Broader Implications}

% Our work contributes to the growing field of visualization for language models, with several broader implications:

% \textbf{Advancing interpretability beyond attention visualization.} While attention visualizations have been widely used to explain language model behavior, our approach demonstrates the value of more granular feature-level analysis. By enabling users to isolate and manipulate specific components, we provide a more mechanistic understanding of model function, complementing the higher-level insights from attention-based approaches.

% \textbf{Supporting model safety and alignment.} Feature-level interpretability tools can play a crucial role in identifying and addressing undesirable model behaviors. By enabling researchers to isolate specific features related to particular capabilities or biases, our approach could support efforts to align models with human values and safety requirements.

% \textbf{Bridging the gap between AI researchers and domain experts.} Our user study suggests that interpretability tools can make language model analysis accessible to a broader audience, including those without specialized ML expertise. This democratization of AI understanding could lead to more diverse perspectives in AI development and more effective application of language models across domains.

In this section, we discuss insights gathered from our user study, the limitations of our system, and directions for future work.

\subsection{Lessons Learned}

Based on the results of our study, we identified several key insights about feature interpretation in large language models:

% 高效的推荐方法降低用户认知负担
\vspace{4pt}
\noindent
% \textbf{Bridging the semantic gap between feature activation and explanation.} 
\textbf{Fusing Automated and Activation-Based Analyses for Robust Feature Interpretation.}
Our study revealed that users highly value tools that help reconcile discrepancies between feature activations and feature explanations. The ability to examine anomalous samples where activation values significantly diverge from similarity provides crucial insights into feature behavior. This suggests the importance of incorporating multiple perspectives when interpreting model components, rather than relying solely on single explanations or activation patterns.

% 帮助用户发现关联特征分析让用户辅助用户理解
\vspace{4pt}
\noindent
% \textbf{Multi-level validation is essential for feature understanding.} 
% Participants consistently emphasized the value of both input and output validation approaches. Input validation helped verify feature-concept relationships by examining activation patterns across different inputs, while output validation demonstrated concrete impacts on model generations. This bi-directional analysis approach allows users to build more complete models of feature function by connecting low-level mechanics to high-level behaviors.
\textbf{Bridging Low-Level Mechanics and High-Level Behaviors for Holistic Feature Analysis.}
Participants consistently emphasized the value of analyzing features from two complementary perspectives. 
They validated features at the level of low-level mechanics through input validation, which helped verify feature-concept relationships by examining internal activation patterns. 
Concurrently, they assessed the high-level behaviors of features through output validation, which demonstrated their concrete impact on the final model generations. 
This bi-directional approach, which effectively bridges these mechanics and behaviors, allowed users to perform a more holistic feature analysis and build a complete mental model of a feature interpretation.

% engagement
\vspace{4pt}
\noindent
% \textbf{Interactive exploration enhances discovery and engagement.} 
\textbf{Facilitating User-Driven Discovery and Engagement through Interactive Exploration.}
Most participants mentioned that the exploration process was engaging and enjoyable. The ability to fluidly move between selecting, browsing, and validation encouraged serendipitous discovery of unexpected feature properties. The interactive nature of our system supported diverse exploration styles: some users preferred a more structured approach starting from concept queries, while others found value in browsing the feature space more freely and following connections between related features. This flexibility allowed users to adapt the system to their own investigative preferences, leading to higher engagement. Many participants reported instances where they discovered novel feature behaviors and unexpected insights they were not actively seeking, highlighting how the interactive design facilitated unexpected insights about model internals.

\subsection{{Limitation and Future Work}}
Despite the positive feedback from users, our system has several limitations that represent opportunities for future research:

% \textcolor{gray}{\textbf{Review:} Given this foundation, it should be clearly highlighted in the limitations section that the tool’s effectiveness depends entirely on the robustness of the SAE method itself, which underpins the generation of mono-semantic features. }
\vspace{4pt}
\noindent
\wz{\textbf{Reliability.}
The reliability of ConceptViz relies on the robustness of underlying SAEs and their automated explanations.
We design our system with this in mind, aiming to help users navigate potential imperfections.
Instead of requiring users to blindly trust these automated outputs, ConceptViz encourages a workflow of exploration and verification, empowering users to confirm or challenge initial findings through direct interaction, thereby building a more robust and evidence-based understanding.
Furthermore, as the research community makes continuous progress in developing more powerful SAEs and explanation techniques, these advancements can be seamlessly integrated into our system, further enhancing its reliability and ease of use.}
% While our system provides tools for feature verification, users still encounter challenges when automated explanations fail to capture feature behavior accurately. Participants valued visual tools that help them intuitively identify where and why automated explanations fall short more than theoretical improvements to automated explanation methods. They suggested enhanced visualization techniques that could directly highlight inconsistencies between a feature's purported explanation and its actual behavior across different inputs. For example, future implementations could automatically flag examples where activation patterns deviate significantly from what would be expected based on the feature explanation or provide comparative visualizations showing the distribution of inputs that match or contradict the expected behavior. These approaches would enable users to quickly identify limitations in feature explanations without requiring exhaustive manual inspection.

\vspace{4pt}
\noindent
\wz{\textbf{Scalability.}
Our current implementation was tested on SAEs with $16,384$ features for each residual layer of the language model. 
As feature spaces scale towards hundreds of thousands or more, the primary computational cost lies in UMAP dimension reduction during data preprocessing. 
In contrast, the interactive components, including vector database queries and frontend rendering, still remain responsive and fluid when tested with up to one million features. 
To efficiently support much larger scales, future work should explore methods like distributed dimension reduction for preprocessing, while further enhancing interactive performance through multi-GPU parallel processing for database queries and adaptive, hierarchical rendering strategies.}
% Our current implementation focuses on specific layers of sparse autoencoders with several thousand features. 
% As language models continue to grow in size and complexity, interpretability systems will need to scale accordingly. 
% Current visualization and navigation approaches become unwieldy when dealing with tens or hundreds of thousands of features.
% Future work should explore methods for navigating much larger feature spaces efficiently, potentially incorporating hierarchical browsing approaches and more sophisticated dimensionality reduction techniques.

\vspace{4pt}
\noindent
\wz{\textbf{Generalizability.}
Our pipeline operates on the embeddings of feature explanations from SAEs, making it independent of any specific LLM architecture. 
Although the case studies in this paper utilize the Gemma-2 model, we have also applied our approach to SAEs trained on GPT-series models. 
This demonstrates the potential to extend ConceptViz to a wider range of LLMs and various emerging SAE architectures. 
Moreover, as multimodal large language models become increasingly prevalent, an exciting direction for future research is to adapt our system for cross-modal feature analysis~\cite{Rao2024Discover,lim2025sparse}, exploring how concepts are represented across modalities like text and images.} 
% As multimodal language models become increasingly prevalent, interpretability tools will need to address cross-modal feature~\cite{Rao2024Discover} interactions. Extending our approach to analyze how features interact across text, vision, and other modalities represents an important direction for future research.
% Our current system is primarily designed as an analysis tool for pretrained models. However, feature interpretability has significant potential for model development and improvement. Several participants with experience in model development noted that insights gained through our system could inform targeted interventions in model architecture or training. Future work could explore tighter integration with model training and refinement workflows, allowing insights gained from feature analysis to directly inform model adjustments. This might include capabilities for marking problematic features, logging insights for development teams, or even interfaces for modifying feature weights to test potential improvements. Such integration would close the loop between analysis and development, making interpretability a more central part of the model improvement process rather than just a post-hoc analysis tool.

\vspace{4pt}
\noindent
\wz{\textbf{Experiment Setting.}
Our experiment setting is constrained by the participant pool, which consisted solely of users with prior SAE knowledge. 
Consequently, our usability findings may not fully capture the challenges faced by novices. 
Meanwhile, the guided tasks were designed to demonstrate core functionality and may not fully represent the complexities of open-ended, real-world research. 
Future work should therefore engage a more diverse user base in more ecologically valid analytical tasks to better assess the system's effectiveness in practice.} 

\section{Conclusion}
This paper presents a visual analytics system, ConceptViz, for exploring and interpreting SAE features in large language models. 
The system integrates identification, interpretation, and validation capabilities in a unified workflow, enabling users to effectively discover, examine, and verify feature behaviors across different contexts. 
Through two usage scenarios and a comprehensive user study, we demonstrate that ConceptViz successfully bridges the gap between automated feature explanations and user understanding. 
By making feature-level analysis more accessible and intuitive, our work contributes to the broader effort of opening the black box of large language models and supports advancements in model interpretability, safety, and alignment. 

\acknowledgments{
We thank anonymous reviewers for their insightful reviews. We also thank Haoyu Tian and Zhaorui Yang for their heartwarming support. This work is supported by the National Natural Science Foundation of China (62132017, 62421003, 62302435), Zhejiang Provincial Natural Science Foundation of China (LD24F020011), and "Pioneer" and "Leading Goose" R\&D Program of Zhejiang (2024C01167).}

\bibliographystyle{abbrv-doi-hyperref}

\bibliography{template}

\begin{thebibliography}{10}

\bibitem{openai2024gpt4}
J.~e.~a. Achiam.
\newblock {GPT-4} technical report.
\newblock {\em arXiv preprint}, 2023. \href{https://doi.org/10.48550/arXiv.2303.08774}
{doi: {{%
10\hspace{.1pt}\discretionary{.}{%
}{.}\hspace{.4pt}48550\discretionary{/}{%
}{/}arXiv\hspace{.1pt}\discretionary{.}{%
}{.}\hspace{.4pt}2303\hspace{.1pt}\discretionary{.}{%
}{.}\hspace{.4pt}08774}}}


\bibitem{TheC3}
Anthropic.
\newblock The claude 3 model family: Opus, sonnet, haiku.
\newblock \url{https://www.anthropic.com/}, 2024.
\newblock Accessed July 2025.

\bibitem{atanasova-etal-2020-diagnostic}
P.~Atanasova, J.~G. Simonsen, C.~Lioma, and I.~Augenstein.
\newblock A diagnostic study of explainability techniques for text classification.
\newblock In {\em Proc. {EMNLP}}, pp. 3256--3274. Association for Computational Linguistics, Stroudsburg, 2020. \href{https://doi.org/10.18653/V1/2020.EMNLP-MAIN.263}
{doi: {{%
10\hspace{.1pt}\discretionary{.}{%
}{.}\hspace{.4pt}18653\discretionary{/}{%
}{/}V1\discretionary{/}{%
}{/}2020\hspace{.1pt}\discretionary{.}{%
}{.}\hspace{.4pt}EMNLP\discretionary{%
}{-}{-}MAIN\hspace{.1pt}\discretionary{.}{%
}{.}\hspace{.4pt}263}}}


\bibitem{balcells2024evolutionsaefeatureslayers}
D.~Balcells, B.~Lerner, M.~Oesterle, E.~Ucar, and S.~Heimersheim.
\newblock Evolution of sae features across layers in llms, 2024.

\bibitem{bills2023language}
S.~Bills, N.~Cammarata, D.~Mossing, H.~Tillman, L.~Gao, G.~Goh, I.~Sutskever, J.~Leike, J.~Wu, and W.~Saunders.
\newblock Language models can explain neurons in language models.
\newblock \url{https://openaipublic.blob.core.windows.net/neuron-explainer/paper/index.html}, 2023.

\bibitem{bloom2024saetrainingcodebase}
J.~Bloom, C.~Tigges, A.~Duong, and D.~Chanin.
\newblock Saelens.
\newblock \url{https://github.com/jbloomAus/SAELens}, 2024.

\bibitem{boggust2025compress}
A.~Boggust, V.~Sivaraman, Y.~Assogba, D.~Ren, D.~Moritz, and F.~Hohman.
\newblock Compress and compare: Interactively evaluating efficiency and behavior across {ML} model compression experiments.
\newblock {\em {IEEE} Trans. Vis. Comput. Graph.}, 31(1):809--819, 2025. \href{https://doi.org/10.1109/TVCG.2024.3456371}
{doi: {{%
10\hspace{.1pt}\discretionary{.}{%
}{.}\hspace{.4pt}1109\discretionary{/}{%
}{/}TVCG\hspace{.1pt}\discretionary{.}{%
}{.}\hspace{.4pt}2024\hspace{.1pt}\discretionary{.}{%
}{.}\hspace{.4pt}3456371}}}


\bibitem{DBLP:journals/corr/abs-2409-14507}
D.~Chanin, J.~Wilken{-}Smith, T.~Dulka, H.~Bhatnagar, and J.~Bloom.
\newblock A is for absorption: Studying feature splitting and absorption in sparse autoencoders.
\newblock {\em arXiv preprint}, abs/2409.14507, 2024. \href{https://doi.org/10.48550/ARXIV.2409.14507}
{doi: {{%
10\hspace{.1pt}\discretionary{.}{%
}{.}\hspace{.4pt}48550\discretionary{/}{%
}{/}ARXIV\hspace{.1pt}\discretionary{.}{%
}{.}\hspace{.4pt}2409\hspace{.1pt}\discretionary{.}{%
}{.}\hspace{.4pt}14507}}}


\bibitem{chaudhary2024evaluatingopensourcesparseautoencoders}
M.~Chaudhary and A.~Geiger.
\newblock Evaluating open-source sparse autoencoders on disentangling factual knowledge in {GPT-2} small.
\newblock {\em arXiv preprint}, abs/2409.04478, 2024. \href{https://doi.org/10.48550/ARXIV.2409.04478}
{doi: {{%
10\hspace{.1pt}\discretionary{.}{%
}{.}\hspace{.4pt}48550\discretionary{/}{%
}{/}ARXIV\hspace{.1pt}\discretionary{.}{%
}{.}\hspace{.4pt}2409\hspace{.1pt}\discretionary{.}{%
}{.}\hspace{.4pt}04478}}}


\bibitem{10669934}
J.~Chen, W.~Yang, Z.~Jia, L.~Xiao, and S.~Liu.
\newblock Dynamic color assignment for hierarchical data.
\newblock {\em {IEEE} Trans. Vis. Comput. Graph.}, 31(1):338--348, 2025. \href{https://doi.org/10.1109/TVCG.2024.3456386}
{doi: {{%
10\hspace{.1pt}\discretionary{.}{%
}{.}\hspace{.4pt}1109\discretionary{/}{%
}{/}TVCG\hspace{.1pt}\discretionary{.}{%
}{.}\hspace{.4pt}2024\hspace{.1pt}\discretionary{.}{%
}{.}\hspace{.4pt}3456386}}}


\bibitem{cho2024transformer}
A.~Cho, G.~C. Kim, A.~Karpekov, A.~Helbling, Z.~J. Wang, S.~Lee, B.~Hoover, and D.~H. Chau.
\newblock Transformer explainer: Interactive learning of text-generative models.
\newblock {\em arXiv preprint}, abs/2408.04619, 2024. \href{https://doi.org/10.48550/ARXIV.2408.04619}
{doi: {{%
10\hspace{.1pt}\discretionary{.}{%
}{.}\hspace{.4pt}48550\discretionary{/}{%
}{/}ARXIV\hspace{.1pt}\discretionary{.}{%
}{.}\hspace{.4pt}2408\hspace{.1pt}\discretionary{.}{%
}{.}\hspace{.4pt}04619}}}


\bibitem{clark-etal-2019-bert}
K.~Clark, U.~Khandelwal, O.~Levy, and C.~D. Manning.
\newblock What does {BERT} look at? an analysis of bert's attention.
\newblock In {\em Proc. BlackboxNLP@ACL}, pp. 276--286. Association for Computational Linguistics, Stroudsburg, 2019. \href{https://doi.org/10.18653/V1/W19-4828}
{doi: {{%
10\hspace{.1pt}\discretionary{.}{%
}{.}\hspace{.4pt}18653\discretionary{/}{%
}{/}V1\discretionary{/}{%
}{/}W19\discretionary{%
}{-}{-}4828}}}


\bibitem{Coscia:2023:KnowledgeVIS}
A.~Coscia and A.~Endert.
\newblock Knowledgevis: Interpreting language models by comparing fill-in-the-blank prompts.
\newblock {\em {IEEE} Trans. Vis. Comput. Graph.}, 30(9):6520--6532, 2024. \href{https://doi.org/10.1109/TVCG.2023.3346713}
{doi: {{%
10\hspace{.1pt}\discretionary{.}{%
}{.}\hspace{.4pt}1109\discretionary{/}{%
}{/}TVCG\hspace{.1pt}\discretionary{.}{%
}{.}\hspace{.4pt}2023\hspace{.1pt}\discretionary{.}{%
}{.}\hspace{.4pt}3346713}}}


\bibitem{Coscia_2024}
A.~Coscia, L.~Holmes, W.~Morris, J.~S. Choi, S.~A. Crossley, and A.~Endert.
\newblock iscore: Visual analytics for interpreting how language models automatically score summaries.
\newblock In {\em Proc. {IUI}}, pp. 787--802. {ACM}, New York, 2024. \href{https://doi.org/10.1145/3640543.3645142}
{doi: {{%
10\hspace{.1pt}\discretionary{.}{%
}{.}\hspace{.4pt}1145\discretionary{/}{%
}{/}3640543\hspace{.1pt}\discretionary{.}{%
}{.}\hspace{.4pt}3645142}}}


\bibitem{DeRose2020AttentionFA}
J.~F. DeRose, J.~Wang, and M.~Berger.
\newblock Attention flows: Analyzing and comparing attention mechanisms in language models.
\newblock {\em {IEEE} Trans. Vis. Comput. Graph.}, 27(2):1160--1170, 2021. \href{https://doi.org/10.1109/TVCG.2020.3028976}
{doi: {{%
10\hspace{.1pt}\discretionary{.}{%
}{.}\hspace{.4pt}1109\discretionary{/}{%
}{/}TVCG\hspace{.1pt}\discretionary{.}{%
}{.}\hspace{.4pt}2020\hspace{.1pt}\discretionary{.}{%
}{.}\hspace{.4pt}3028976}}}


\bibitem{elhage2022superposition}
N.~Elhage, T.~Hume, C.~Olsson, N.~Schiefer, T.~Henighan, S.~Kravec, Z.~Hatfield-Dodds, R.~Lasenby, D.~Drain, C.~Chen, R.~Grosse, S.~McCandlish, J.~Kaplan, D.~Amodei, M.~Wattenberg, and C.~Olah.
\newblock Toy models of superposition.
\newblock {\em arXiv preprint}, abs/2209.10652, 2022. \href{https://doi.org/10.48550/ARXIV.2209.10652}
{doi: {{%
10\hspace{.1pt}\discretionary{.}{%
}{.}\hspace{.4pt}48550\discretionary{/}{%
}{/}ARXIV\hspace{.1pt}\discretionary{.}{%
}{.}\hspace{.4pt}2209\hspace{.1pt}\discretionary{.}{%
}{.}\hspace{.4pt}10652}}}


\bibitem{templeton2024scaling}
A.~T. et~al.
\newblock Scaling monosemanticity: Extracting interpretable features from claude 3 sonnet.
\newblock {\em Transformer Circuits Thread}, 2024.

\bibitem{olsson2022context}
C.~O. et~al.
\newblock In-context learning and induction heads.
\newblock {\em arXiv preprint}, abs/2209.11895, 2022. \href{https://doi.org/10.48550/ARXIV.2209.11895}
{doi: {{%
10\hspace{.1pt}\discretionary{.}{%
}{.}\hspace{.4pt}48550\discretionary{/}{%
}{/}ARXIV\hspace{.1pt}\discretionary{.}{%
}{.}\hspace{.4pt}2209\hspace{.1pt}\discretionary{.}{%
}{.}\hspace{.4pt}11895}}}


\bibitem{durmus2024steering}
E.~D. et~al.
\newblock Evaluating feature steering: A case study in mitigating social biases.
\newblock \url{https://www.anthropic.com/research/evaluating-feature-steering}, 2024.
\newblock Accessed July 2025.

\bibitem{gemmateam2024gemma2}
G.~T. et~al.
\newblock Gemma 2: Improving open language models at a practical size.
\newblock {\em arXiv preprint}, abs/2408.00118, 2024. \href{https://doi.org/10.48550/ARXIV.2408.00118}
{doi: {{%
10\hspace{.1pt}\discretionary{.}{%
}{.}\hspace{.4pt}48550\discretionary{/}{%
}{/}ARXIV\hspace{.1pt}\discretionary{.}{%
}{.}\hspace{.4pt}2408\hspace{.1pt}\discretionary{.}{%
}{.}\hspace{.4pt}00118}}}


\bibitem{kahng2025comparator}
M.~K. et~al.
\newblock {LLM} comparator: Interactive analysis of side-by-side evaluation of large language models.
\newblock {\em {IEEE} Trans. Vis. Comput. Graph.}, 31(1):503--513, 2025. \href{https://doi.org/10.1109/TVCG.2024.3456354}
{doi: {{%
10\hspace{.1pt}\discretionary{.}{%
}{.}\hspace{.4pt}1109\discretionary{/}{%
}{/}TVCG\hspace{.1pt}\discretionary{.}{%
}{.}\hspace{.4pt}2024\hspace{.1pt}\discretionary{.}{%
}{.}\hspace{.4pt}3456354}}}


\bibitem{bricken2023monosemanticity}
T.~B. et~al.
\newblock Towards monosemanticity: Decomposing language models with dictionary learning.
\newblock \url{https://transformer-circuits.pub/2023/monosemantic-features/index.html}, 2023.
\newblock Accessed July 2025.

\bibitem{he2024llamascopeextractingmillions}
Z.~H. et~al.
\newblock Llama scope: Extracting millions of features from llama-3.1-8b with sparse autoencoders.
\newblock {\em arXiv preprint}, abs/2410.20526, 2024. \href{https://doi.org/10.48550/ARXIV.2410.20526}
{doi: {{%
10\hspace{.1pt}\discretionary{.}{%
}{.}\hspace{.4pt}48550\discretionary{/}{%
}{/}ARXIV\hspace{.1pt}\discretionary{.}{%
}{.}\hspace{.4pt}2410\hspace{.1pt}\discretionary{.}{%
}{.}\hspace{.4pt}20526}}}


\bibitem{DBLP:journals/corr/abs-2410-19278}
E.~Farrell, Y.~Lau, and A.~Conmy.
\newblock Applying sparse autoencoders to unlearn knowledge in language models.
\newblock {\em arXiv preprint}, abs/2410.19278, 2024. \href{https://doi.org/10.48550/ARXIV.2410.19278}
{doi: {{%
10\hspace{.1pt}\discretionary{.}{%
}{.}\hspace{.4pt}48550\discretionary{/}{%
}{/}ARXIV\hspace{.1pt}\discretionary{.}{%
}{.}\hspace{.4pt}2410\hspace{.1pt}\discretionary{.}{%
}{.}\hspace{.4pt}19278}}}


\bibitem{PromptMagician}
Y.~Feng, X.~Wang, K.~K. Wong, S.~Wang, Y.~Lu, M.~Zhu, B.~Wang, and W.~Chen.
\newblock { PromptMagician: Interactive Prompt Engineering for Text-to-Image Creation }.
\newblock {\em {IEEE} Trans. Vis. Comput. Graph.}, 30(01):295--305, Jan. 2024. \href{https://doi.org/10.1109/TVCG.2023.3327168}
{doi: {{%
10\hspace{.1pt}\discretionary{.}{%
}{.}\hspace{.4pt}1109\discretionary{/}{%
}{/}TVCG\hspace{.1pt}\discretionary{.}{%
}{.}\hspace{.4pt}2023\hspace{.1pt}\discretionary{.}{%
}{.}\hspace{.4pt}3327168}}}


\bibitem{ferrando2024information}
J.~Ferrando and E.~Voita.
\newblock Information flow routes: Automatically interpreting language models at scale.
\newblock In {\em Proc. {EMNLP}}, pp. 17432--17445. Association for Computational Linguistics, Miami, 2024. \href{https://doi.org/10.18653/v1/2024.emnlp-main.965}
{doi: {{%
10\hspace{.1pt}\discretionary{.}{%
}{.}\hspace{.4pt}18653\discretionary{/}{%
}{/}v1\discretionary{/}{%
}{/}2024\hspace{.1pt}\discretionary{.}{%
}{.}\hspace{.4pt}emnlp\discretionary{%
}{-}{-}main\hspace{.1pt}\discretionary{.}{%
}{.}\hspace{.4pt}965}}}


\bibitem{foote2023ng}
A.~Foote, N.~Nanda, E.~Kran, I.~Konstas, and F.~Barez.
\newblock {N2G:} {A} scalable approach for quantifying interpretable neuron representations in large language models.
\newblock {\em arXiv preprint}, abs/2304.12918, 2023. \href{https://doi.org/10.48550/ARXIV.2304.12918}
{doi: {{%
10\hspace{.1pt}\discretionary{.}{%
}{.}\hspace{.4pt}48550\discretionary{/}{%
}{/}ARXIV\hspace{.1pt}\discretionary{.}{%
}{.}\hspace{.4pt}2304\hspace{.1pt}\discretionary{.}{%
}{.}\hspace{.4pt}12918}}}


\bibitem{gao2024scaling}
L.~Gao, T.~D. la~Tour, H.~Tillman, G.~Goh, R.~Troll, A.~Radford, I.~Sutskever, J.~Leike, and J.~Wu.
\newblock Scaling and evaluating sparse autoencoders.
\newblock {\em arXiv preprint}, abs/2406.04093, 2024. \href{https://doi.org/10.48550/ARXIV.2406.04093}
{doi: {{%
10\hspace{.1pt}\discretionary{.}{%
}{.}\hspace{.4pt}48550\discretionary{/}{%
}{/}ARXIV\hspace{.1pt}\discretionary{.}{%
}{.}\hspace{.4pt}2406\hspace{.1pt}\discretionary{.}{%
}{.}\hspace{.4pt}04093}}}


\bibitem{10297598}
L.~Gao, Z.~Shao, Z.~Luo, H.~Hu, C.~Turkay, and S.~Chen.
\newblock Transforlearn: Interactive visual tutorial for the transformer model.
\newblock {\em {IEEE} Trans. Vis. Comput. Graph.}, 30(1):891--901, 2024. \href{https://doi.org/10.1109/TVCG.2023.3327353}
{doi: {{%
10\hspace{.1pt}\discretionary{.}{%
}{.}\hspace{.4pt}1109\discretionary{/}{%
}{/}TVCG\hspace{.1pt}\discretionary{.}{%
}{.}\hspace{.4pt}2023\hspace{.1pt}\discretionary{.}{%
}{.}\hspace{.4pt}3327353}}}


\bibitem{8371286}
F.~Hohman, M.~Kahng, R.~S. Pienta, and D.~H. Chau.
\newblock Visual analytics in deep learning: An interrogative survey for the next frontiers.
\newblock {\em {IEEE} Trans. Vis. Comput. Graph.}, 25(8):2674--2693, 2019. \href{https://doi.org/10.1109/TVCG.2018.2843369}
{doi: {{%
10\hspace{.1pt}\discretionary{.}{%
}{.}\hspace{.4pt}1109\discretionary{/}{%
}{/}TVCG\hspace{.1pt}\discretionary{.}{%
}{.}\hspace{.4pt}2018\hspace{.1pt}\discretionary{.}{%
}{.}\hspace{.4pt}2843369}}}


\bibitem{VisualConceptProgramming}
M.~N. Hoque, W.~He, A.~K. Shekar, L.~Gou, and L.~Ren.
\newblock { Visual Concept Programming: A Visual Analytics Approach to Injecting Human Intelligence at Scale }.
\newblock {\em {IEEE} Trans. Vis. Comput. Graph.}, 29(01):74--83, Jan. 2023. \href{https://doi.org/10.1109/TVCG.2022.3209466}
{doi: {{%
10\hspace{.1pt}\discretionary{.}{%
}{.}\hspace{.4pt}1109\discretionary{/}{%
}{/}TVCG\hspace{.1pt}\discretionary{.}{%
}{.}\hspace{.4pt}2022\hspace{.1pt}\discretionary{.}{%
}{.}\hspace{.4pt}3209466}}}


\bibitem{ConceptExplainer}
J.~Huang, A.~Mishra, B.~C. Kwon, and C.~Bryan.
\newblock { ConceptExplainer: Interactive Explanation for Deep Neural Networks from a Concept Perspective }.
\newblock {\em {IEEE} Trans. Vis. Comput. Graph.}, 29(01):831--841, Jan. 2023. \href{https://doi.org/10.1109/TVCG.2022.3209384}
{doi: {{%
10\hspace{.1pt}\discretionary{.}{%
}{.}\hspace{.4pt}1109\discretionary{/}{%
}{/}TVCG\hspace{.1pt}\discretionary{.}{%
}{.}\hspace{.4pt}2022\hspace{.1pt}\discretionary{.}{%
}{.}\hspace{.4pt}3209384}}}


\bibitem{cunningham2023sparseautoencodershighlyinterpretable}
R.~Huben, H.~Cunningham, L.~R. Smith, A.~Ewart, and L.~Sharkey.
\newblock Sparse autoencoders find highly interpretable features in language models.
\newblock In {\em {ICLR}}. OpenReview.net, Vienna, 2024. \href{https://doi.org/10.48550/ARXIV.2309.08600}
{doi: {{%
10\hspace{.1pt}\discretionary{.}{%
}{.}\hspace{.4pt}48550\discretionary{/}{%
}{/}ARXIV\hspace{.1pt}\discretionary{.}{%
}{.}\hspace{.4pt}2309\hspace{.1pt}\discretionary{.}{%
}{.}\hspace{.4pt}08600}}}


\bibitem{karvonen2025saebench}
A.~Karvonen, C.~Rager, J.~Lin, C.~Tigges, J.~I. Bloom, D.~Chanin, Y.-T. Lau, E.~Farrell, C.~McDougall, K.~Ayonrinde, M.~Wearden, A.~Conmy, S.~Marks, and N.~Nanda.
\newblock Saebench: A comprehensive benchmark for sparse autoencoders in language model interpretability.
\newblock {\em arXiv preprint}, abs/2503.09532, 2025. \href{https://doi.org/10.48550/arXiv.2503.09532}
{doi: {{%
10\hspace{.1pt}\discretionary{.}{%
}{.}\hspace{.4pt}48550\discretionary{/}{%
}{/}arXiv\hspace{.1pt}\discretionary{.}{%
}{.}\hspace{.4pt}2503\hspace{.1pt}\discretionary{.}{%
}{.}\hspace{.4pt}09532}}}


\bibitem{karvonen2025measuringprogressdictionarylearning}
A.~Karvonen, B.~Wright, C.~Rager, R.~Angell, J.~Brinkmann, L.~Smith, C.~Mayrink~Verdun, D.~Bau, and S.~Marks.
\newblock Measuring progress in dictionary learning for language model interpretability with board game models.
\newblock In A.~Globerson, L.~Mackey, D.~Belgrave, A.~Fan, U.~Paquet, J.~Tomczak, and C.~Zhang, eds., {\em Proc. NeurIPS}, vol.~37, pp. 83091--83118. Curran Associates, Inc., 2024. \href{https://doi.org/10.48550/arXiv.2408.00113}
{doi: {{%
10\hspace{.1pt}\discretionary{.}{%
}{.}\hspace{.4pt}48550\discretionary{/}{%
}{/}arXiv\hspace{.1pt}\discretionary{.}{%
}{.}\hspace{.4pt}2408\hspace{.1pt}\discretionary{.}{%
}{.}\hspace{.4pt}00113}}}


\bibitem{lieberum-etal-2024-gemma}
T.~Lieberum, S.~Rajamanoharan, A.~Conmy, L.~Smith, N.~Sonnerat, V.~Varma, J.~Kram{\'{a}}r, A.~D. Dragan, R.~Shah, and N.~Nanda.
\newblock Gemma scope: Open sparse autoencoders everywhere all at once on gemma 2.
\newblock {\em arXiv preprint}, abs/2408.05147, 2024. \href{https://doi.org/10.48550/ARXIV.2408.05147}
{doi: {{%
10\hspace{.1pt}\discretionary{.}{%
}{.}\hspace{.4pt}48550\discretionary{/}{%
}{/}ARXIV\hspace{.1pt}\discretionary{.}{%
}{.}\hspace{.4pt}2408\hspace{.1pt}\discretionary{.}{%
}{.}\hspace{.4pt}05147}}}


\bibitem{lim2025sparse}
H.~Lim, J.~Choi, J.~Choo, and S.~Schneider.
\newblock Sparse autoencoders reveal selective remapping of visual concepts during adaptation.
\newblock In {\em Proc. ICLR}. OpenReview.net, Singapore, 2025. \href{https://doi.org/10.48550/arXiv.2412.05276}
{doi: {{%
10\hspace{.1pt}\discretionary{.}{%
}{.}\hspace{.4pt}48550\discretionary{/}{%
}{/}arXiv\hspace{.1pt}\discretionary{.}{%
}{.}\hspace{.4pt}2412\hspace{.1pt}\discretionary{.}{%
}{.}\hspace{.4pt}05276}}}


\bibitem{article}
S.~Liu, X.~Wang, M.~Liu, and J.~Zhu.
\newblock Towards better analysis of machine learning models: {A} visual analytics perspective.
\newblock {\em Vis. Informatics}, 1(1):48--56, 2017. \href{https://doi.org/10.1016/J.VISINF.2017.01.006}
{doi: {{%
10\hspace{.1pt}\discretionary{.}{%
}{.}\hspace{.4pt}1016\discretionary{/}{%
}{/}J\hspace{.1pt}\discretionary{.}{%
}{.}\hspace{.4pt}VISINF\hspace{.1pt}\discretionary{.}{%
}{.}\hspace{.4pt}2017\hspace{.1pt}\discretionary{.}{%
}{.}\hspace{.4pt}01\hspace{.1pt}\discretionary{.}{%
}{.}\hspace{.4pt}006}}}


\bibitem{makelov2024sparse}
A.~Makelov.
\newblock Sparse autoencoders match supervised features for model steering on the {IOI} task.
\newblock In {\em Proc. ICML 2024 Workshop on Mechanistic Interpretability}. ICML, San Diego, 2024.

\bibitem{marks2024sparsefeaturecircuitsdiscovering}
S.~Marks, C.~Rager, E.~J. Michaud, Y.~Belinkov, D.~Bau, and A.~Mueller.
\newblock Sparse feature circuits: Discovering and editing interpretable causal graphs in language models.
\newblock {\em arXiv preprint}, abs/2403.19647, 2024. \href{https://doi.org/10.48550/ARXIV.2403.19647}
{doi: {{%
10\hspace{.1pt}\discretionary{.}{%
}{.}\hspace{.4pt}48550\discretionary{/}{%
}{/}ARXIV\hspace{.1pt}\discretionary{.}{%
}{.}\hspace{.4pt}2403\hspace{.1pt}\discretionary{.}{%
}{.}\hspace{.4pt}19647}}}


\bibitem{marks2024feature}
S.~Marks, C.~Rager, E.~J. Michaud, Y.~Belinkov, D.~Bau, and A.~Mueller.
\newblock Sparse feature circuits: Discovering and editing interpretable causal graphs in language models.
\newblock {\em arXiv preprint}, abs/2403.19647, 2024. \href{https://doi.org/10.48550/ARXIV.2403.19647}
{doi: {{%
10\hspace{.1pt}\discretionary{.}{%
}{.}\hspace{.4pt}48550\discretionary{/}{%
}{/}ARXIV\hspace{.1pt}\discretionary{.}{%
}{.}\hspace{.4pt}2403\hspace{.1pt}\discretionary{.}{%
}{.}\hspace{.4pt}19647}}}


\bibitem{meng2022locating}
K.~Meng, D.~Bau, A.~Andonian, and Y.~Belinkov.
\newblock Locating and editing factual associations in gpt.
\newblock In S.~Koyejo, S.~Mohamed, A.~Agarwal, D.~Belgrave, K.~Cho, and A.~Oh, eds., {\em Proc. NeurIPS}, vol.~35, pp. 17359--17372. Curran Associates, Inc., New Orleans, 2022. \href{https://doi.org/10.48550/arXiv.2202.05262}
{doi: {{%
10\hspace{.1pt}\discretionary{.}{%
}{.}\hspace{.4pt}48550\discretionary{/}{%
}{/}arXiv\hspace{.1pt}\discretionary{.}{%
}{.}\hspace{.4pt}2202\hspace{.1pt}\discretionary{.}{%
}{.}\hspace{.4pt}05262}}}


\bibitem{beren2022singular}
B.~Millidge and S.~Black.
\newblock The singular value decompositions of transformer weight matrices are highly interpretable.
\newblock \url{https://bit.ly/3GdbZoa}, 2022.

\bibitem{olah2020zoom}
C.~Olah, N.~Cammarata, L.~Schubert, G.~Goh, M.~Petrov, and S.~Carter.
\newblock Zoom in: An introduction to circuits.
\newblock {\em Distill}, 2020. \href{https://doi.org/10.23915/distill.00024.001}
{doi: {{%
10\hspace{.1pt}\discretionary{.}{%
}{.}\hspace{.4pt}23915\discretionary{/}{%
}{/}distill\hspace{.1pt}\discretionary{.}{%
}{.}\hspace{.4pt}00024\hspace{.1pt}\discretionary{.}{%
}{.}\hspace{.4pt}001}}}


\bibitem{paulo2025automatically}
G.~Paulo, A.~Mallen, C.~Juang, and N.~Belrose.
\newblock Automatically interpreting millions of features in large language models.
\newblock {\em arXiv preprint}, abs/2410.13928, 2024. \href{https://doi.org/10.48550/ARXIV.2410.13928}
{doi: {{%
10\hspace{.1pt}\discretionary{.}{%
}{.}\hspace{.4pt}48550\discretionary{/}{%
}{/}ARXIV\hspace{.1pt}\discretionary{.}{%
}{.}\hspace{.4pt}2410\hspace{.1pt}\discretionary{.}{%
}{.}\hspace{.4pt}13928}}}


\bibitem{rajamanoharan2024improvingdictionarylearninggated}
S.~Rajamanoharan, A.~Conmy, L.~Smith, T.~Lieberum, V.~Varma, J.~Kram{\'{a}}r, R.~Shah, and N.~Nanda.
\newblock Improving dictionary learning with gated sparse autoencoders.
\newblock {\em arXiv preprint}, abs/2404.16014, 2024. \href{https://doi.org/10.48550/ARXIV.2404.16014}
{doi: {{%
10\hspace{.1pt}\discretionary{.}{%
}{.}\hspace{.4pt}48550\discretionary{/}{%
}{/}ARXIV\hspace{.1pt}\discretionary{.}{%
}{.}\hspace{.4pt}2404\hspace{.1pt}\discretionary{.}{%
}{.}\hspace{.4pt}16014}}}


\bibitem{Rajamanoharan2024JumpingAI}
S.~Rajamanoharan, T.~Lieberum, N.~Sonnerat, A.~Conmy, V.~Varma, J.~Kram{\'{a}}r, and N.~Nanda.
\newblock Jumping ahead: Improving reconstruction fidelity with jumprelu sparse autoencoders.
\newblock {\em arXiv preprint}, abs/2407.14435, 2024. \href{https://doi.org/10.48550/ARXIV.2407.14435}
{doi: {{%
10\hspace{.1pt}\discretionary{.}{%
}{.}\hspace{.4pt}48550\discretionary{/}{%
}{/}ARXIV\hspace{.1pt}\discretionary{.}{%
}{.}\hspace{.4pt}2407\hspace{.1pt}\discretionary{.}{%
}{.}\hspace{.4pt}14435}}}


\bibitem{Rao2024Discover}
S.~Rao, S.~Mahajan, M.~B{\"o}hle, and B.~Schiele.
\newblock Discover-then-name: Task-agnostic concept bottlenecks via automated concept discovery.
\newblock In {\em Proc. ECCV}, pp. 444--461. Springer Nature Switzerland, Cham, 2024. \href{https://doi.org/10.1007/978-3-031-72980-5_26}
{doi: {{%
10\hspace{.1pt}\discretionary{.}{%
}{.}\hspace{.4pt}1007\discretionary{/}{%
}{/}978\discretionary{%
}{-}{-}3\discretionary{%
}{-}{-}031\discretionary{%
}{-}{-}72980\discretionary{%
}{-}{-}5\_26}}}


\bibitem{shu2025inputactivationsidentifyinginfluential}
D.~Shu, X.~Wu, H.~Zhao, M.~Du, and N.~Liu.
\newblock Beyond input activations: Identifying influential latents by gradient sparse autoencoders, 2025.

\bibitem{shu2025surveysparseautoencodersinterpreting}
D.~Shu, X.~Wu, H.~Zhao, D.~Rai, Z.~Yao, N.~Liu, and M.~Du.
\newblock A survey on sparse autoencoders: Interpreting the internal mechanisms of large language models, 2025.

\bibitem{8494828}
H.~Strobelt, S.~Gehrmann, M.~Behrisch, A.~Perer, H.~Pfister, and A.~M. Rush.
\newblock Seq2seq-vis: {A} visual debugging tool for sequence-to-sequence models.
\newblock {\em {IEEE} Trans. Vis. Comput. Graph.}, 25(1):353--363, 2019. \href{https://doi.org/10.1109/TVCG.2018.2865044}
{doi: {{%
10\hspace{.1pt}\discretionary{.}{%
}{.}\hspace{.4pt}1109\discretionary{/}{%
}{/}TVCG\hspace{.1pt}\discretionary{.}{%
}{.}\hspace{.4pt}2018\hspace{.1pt}\discretionary{.}{%
}{.}\hspace{.4pt}2865044}}}


\bibitem{8017583}
H.~Strobelt, S.~Gehrmann, H.~Pfister, and A.~M. Rush.
\newblock Lstmvis: {A} tool for visual analysis of hidden state dynamics in recurrent neural networks.
\newblock {\em {IEEE} Trans. Vis. Comput. Graph.}, 24(1):667--676, 2018. \href{https://doi.org/10.1109/TVCG.2017.2744158}
{doi: {{%
10\hspace{.1pt}\discretionary{.}{%
}{.}\hspace{.4pt}1109\discretionary{/}{%
}{/}TVCG\hspace{.1pt}\discretionary{.}{%
}{.}\hspace{.4pt}2017\hspace{.1pt}\discretionary{.}{%
}{.}\hspace{.4pt}2744158}}}


\bibitem{strobelt-etal-2021-lmdiff}
H.~Strobelt, B.~Hoover, A.~Satyanarayan, and S.~Gehrmann.
\newblock Lmdiff: {A} visual diff tool to compare language models.
\newblock In {\em Proc. {EMNLP}}, pp. 96--105. Association for Computational Linguistics, Stroudsburg, 2021. \href{https://doi.org/10.18653/V1/2021.EMNLP-DEMO.12}
{doi: {{%
10\hspace{.1pt}\discretionary{.}{%
}{.}\hspace{.4pt}18653\discretionary{/}{%
}{/}V1\discretionary{/}{%
}{/}2021\hspace{.1pt}\discretionary{.}{%
}{.}\hspace{.4pt}EMNLP\discretionary{%
}{-}{-}DEMO\hspace{.1pt}\discretionary{.}{%
}{.}\hspace{.4pt}12}}}


\bibitem{9908590}
H.~Strobelt, A.~Webson, V.~Sanh, B.~Hoover, J.~Beyer, H.~Pfister, and A.~M. Rush.
\newblock Interactive and visual prompt engineering for ad-hoc task adaptation with large language models.
\newblock {\em {IEEE} Trans. Vis. Comput. Graph.}, 29(1):1146--1156, 2023. \href{https://doi.org/10.1109/TVCG.2022.3209479}
{doi: {{%
10\hspace{.1pt}\discretionary{.}{%
}{.}\hspace{.4pt}1109\discretionary{/}{%
}{/}TVCG\hspace{.1pt}\discretionary{.}{%
}{.}\hspace{.4pt}2022\hspace{.1pt}\discretionary{.}{%
}{.}\hspace{.4pt}3209479}}}


\bibitem{tai-etal-2020-exbert}
W.~Tai, H.~T. Kung, X.~Dong, M.~Z. Comiter, and C.~Kuo.
\newblock exbert: Extending pre-trained models with domain-specific vocabulary under constrained training resources.
\newblock In {\em Proc. {EMNLP}}, vol. {EMNLP} 2020 of {\em Findings of {ACL}}, pp. 1433--1439. ACL, Stroudsburg, 2020. \href{https://doi.org/10.18653/V1/2020.FINDINGS-EMNLP.129}
{doi: {{%
10\hspace{.1pt}\discretionary{.}{%
}{.}\hspace{.4pt}18653\discretionary{/}{%
}{/}V1\discretionary{/}{%
}{/}2020\hspace{.1pt}\discretionary{.}{%
}{.}\hspace{.4pt}FINDINGS\discretionary{%
}{-}{-}EMNLP\hspace{.1pt}\discretionary{.}{%
}{.}\hspace{.4pt}129}}}


\bibitem{tufanov2024lm}
I.~Tufanov, K.~Hambardzumyan, J.~Ferrando, and E.~Voita.
\newblock {LM} transparency tool: Interactive tool for analyzing transformer language models.
\newblock {\em arXiv preprint}, abs/2404.07004, 2024. \href{https://doi.org/10.48550/ARXIV.2404.07004}
{doi: {{%
10\hspace{.1pt}\discretionary{.}{%
}{.}\hspace{.4pt}48550\discretionary{/}{%
}{/}ARXIV\hspace{.1pt}\discretionary{.}{%
}{.}\hspace{.4pt}2404\hspace{.1pt}\discretionary{.}{%
}{.}\hspace{.4pt}07004}}}


\bibitem{DBLP:journals/corr/abs-1906-05714}
J.~Vig.
\newblock A multiscale visualization of attention in the transformer model.
\newblock In {\em Proc. {ACL}}, pp. 37--42. Association for Computational Linguistics, Stroudsburg, 2019. \href{https://doi.org/10.18653/V1/P19-3007}
{doi: {{%
10\hspace{.1pt}\discretionary{.}{%
}{.}\hspace{.4pt}18653\discretionary{/}{%
}{/}V1\discretionary{/}{%
}{/}P19\discretionary{%
}{-}{-}3007}}}


\bibitem{Wang2022InterpretabilityIT}
K.~R. Wang, A.~Variengien, A.~Conmy, B.~Shlegeris, and J.~Steinhardt.
\newblock Interpretability in the wild: a circuit for indirect object identification in {GPT-2} small.
\newblock In {\em {ICLR}}. OpenReview.net, Kigali, 2023. \href{https://doi.org/10.48550/arXiv.2211.00593}
{doi: {{%
10\hspace{.1pt}\discretionary{.}{%
}{.}\hspace{.4pt}48550\discretionary{/}{%
}{/}arXiv\hspace{.1pt}\discretionary{.}{%
}{.}\hspace{.4pt}2211\hspace{.1pt}\discretionary{.}{%
}{.}\hspace{.4pt}00593}}}


\bibitem{DRAVA}
Q.~Wang, S.~L'Yi, and N.~Gehlenborg.
\newblock Drava: Aligning human concepts with machine learning latent dimensions for the visual exploration of small multiples.
\newblock In {\em Proc. CHI}, CHI '23,  article no. 833,  15 pages. Association for Computing Machinery, New York, 2023. \href{https://doi.org/10.1145/3544548.3581127}
{doi: {{%
10\hspace{.1pt}\discretionary{.}{%
}{.}\hspace{.4pt}1145\discretionary{/}{%
}{/}3544548\hspace{.1pt}\discretionary{.}{%
}{.}\hspace{.4pt}3581127}}}


\bibitem{Wang_2023}
X.~Wang, R.~Huang, Z.~Jin, T.~Fang, and H.~Qu.
\newblock Commonsensevis: Visualizing and understanding commonsense reasoning capabilities of natural language models.
\newblock {\em {IEEE} Trans. Vis. Comput. Graph.}, 30(1):273--283, 2024. \href{https://doi.org/10.1109/TVCG.2023.3327153}
{doi: {{%
10\hspace{.1pt}\discretionary{.}{%
}{.}\hspace{.4pt}1109\discretionary{/}{%
}{/}TVCG\hspace{.1pt}\discretionary{.}{%
}{.}\hspace{.4pt}2023\hspace{.1pt}\discretionary{.}{%
}{.}\hspace{.4pt}3327153}}}


\bibitem{wangDodrioExploringTransformer2021}
Z.~J. Wang, R.~Turko, and D.~H. Chau.
\newblock Dodrio: Exploring transformer models with interactive visualization.
\newblock In {\em Proc. {ACL}}, pp. 132--141. Association for Computational Linguistics, Stroudsburg, 2021. \href{https://doi.org/10.18653/V1/2021.ACL-DEMO.16}
{doi: {{%
10\hspace{.1pt}\discretionary{.}{%
}{.}\hspace{.4pt}18653\discretionary{/}{%
}{/}V1\discretionary{/}{%
}{/}2021\hspace{.1pt}\discretionary{.}{%
}{.}\hspace{.4pt}ACL\discretionary{%
}{-}{-}DEMO\hspace{.1pt}\discretionary{.}{%
}{.}\hspace{.4pt}16}}}


\bibitem{wangCNNExplainerLearning2020}
Z.~J. Wang, R.~Turko, O.~Shaikh, H.~Park, N.~Das, F.~Hohman, M.~Kahng, and D.~H.~P. Chau.
\newblock {CNN} explainer: Learning convolutional neural networks with interactive visualization.
\newblock {\em {IEEE} Trans. Vis. Comput. Graph.}, 27(2):1396--1406, 2021. \href{https://doi.org/10.1109/TVCG.2020.3030418}
{doi: {{%
10\hspace{.1pt}\discretionary{.}{%
}{.}\hspace{.4pt}1109\discretionary{/}{%
}{/}TVCG\hspace{.1pt}\discretionary{.}{%
}{.}\hspace{.4pt}2020\hspace{.1pt}\discretionary{.}{%
}{.}\hspace{.4pt}3030418}}}


\bibitem{wendler-etal-2024-llamas}
C.~Wendler, V.~Veselovsky, G.~Monea, and R.~West.
\newblock Do llamas work in english? on the latent language of multilingual transformers.
\newblock In {\em Proc. {ACL}}, pp. 15366--15394. Association for Computational Linguistics, Stroudsburg, 2024. \href{https://doi.org/10.18653/V1/2024.ACL-LONG.820}
{doi: {{%
10\hspace{.1pt}\discretionary{.}{%
}{.}\hspace{.4pt}18653\discretionary{/}{%
}{/}V1\discretionary{/}{%
}{/}2024\hspace{.1pt}\discretionary{.}{%
}{.}\hspace{.4pt}ACL\discretionary{%
}{-}{-}LONG\hspace{.1pt}\discretionary{.}{%
}{.}\hspace{.4pt}820}}}


\bibitem{jain-wallace-2019-attention}
S.~Wiegreffe and Y.~Pinter.
\newblock Attention is not not explanation.
\newblock In {\em Proc. {EMNLP/IJCNLP}}, pp. 11--20. Association for Computational Linguistics, Stroudsburg, 2019. \href{https://doi.org/10.18653/V1/D19-1002}
{doi: {{%
10\hspace{.1pt}\discretionary{.}{%
}{.}\hspace{.4pt}18653\discretionary{/}{%
}{/}V1\discretionary{/}{%
}{/}D19\discretionary{%
}{-}{-}1002}}}


\bibitem{WOODMAN2024101238}
O.~Woodman, Z.~Wen, H.~Lu, Y.~Ren, M.~Zhu, and W.~Chen.
\newblock Exploring the neural landscape: Visual analytics of neuron activation in large language models with neuronautllm.
\newblock {\em Graph. Model.}, 136:101238, 2024. \href{https://doi.org/10.1016/J.GMOD.2024.101238}
{doi: {{%
10\hspace{.1pt}\discretionary{.}{%
}{.}\hspace{.4pt}1016\discretionary{/}{%
}{/}J\hspace{.1pt}\discretionary{.}{%
}{.}\hspace{.4pt}GMOD\hspace{.1pt}\discretionary{.}{%
}{.}\hspace{.4pt}2024\hspace{.1pt}\discretionary{.}{%
}{.}\hspace{.4pt}101238}}}


\bibitem{wu-etal-2024-language}
X.~Wu, W.~Yao, J.~Chen, X.~Pan, X.~Wang, N.~Liu, and D.~Yu.
\newblock From language modeling to instruction following: Understanding the behavior shift in llms after instruction tuning.
\newblock In {\em {NAACL-HLT}}, pp. 2341--2369. Association for Computational Linguistics, Mexico City, 2024. \href{https://doi.org/10.18653/V1/2024.NAACL-LONG.130}
{doi: {{%
10\hspace{.1pt}\discretionary{.}{%
}{.}\hspace{.4pt}18653\discretionary{/}{%
}{/}V1\discretionary{/}{%
}{/}2024\hspace{.1pt}\discretionary{.}{%
}{.}\hspace{.4pt}NAACL\discretionary{%
}{-}{-}LONG\hspace{.1pt}\discretionary{.}{%
}{.}\hspace{.4pt}130}}}


\bibitem{wu2025interpretingsteeringllmsmutual}
X.~Wu, J.~Yuan, W.~Yao, X.~Zhai, and N.~Liu.
\newblock Interpreting and steering llms with mutual information-based explanations on sparse autoencoders.
\newblock {\em arXiv preprint}, abs/2502.15576, 2025. \href{https://doi.org/10.48550/ARXIV.2502.15576}
{doi: {{%
10\hspace{.1pt}\discretionary{.}{%
}{.}\hspace{.4pt}48550\discretionary{/}{%
}{/}ARXIV\hspace{.1pt}\discretionary{.}{%
}{.}\hspace{.4pt}2502\hspace{.1pt}\discretionary{.}{%
}{.}\hspace{.4pt}15576}}}


\bibitem{yang2024foundation}
W.~Yang, M.~Liu, Z.~Wang, and S.~Liu.
\newblock Foundation models meet visualizations: Challenges and opportunities.
\newblock {\em Computational Visual Media}, 10(3):399--424, 2024.

\bibitem{yeh2023attentionviz}
C.~Yeh, Y.~Chen, A.~Wu, C.~Chen, F.~B. Vi{\'{e}}gas, and M.~Wattenberg.
\newblock Attentionviz: {A} global view of transformer attention.
\newblock {\em {IEEE} Trans. Vis. Comput. Graph.}, 30(1):262--272, 2024. \href{https://doi.org/10.1109/TVCG.2023.3327163}
{doi: {{%
10\hspace{.1pt}\discretionary{.}{%
}{.}\hspace{.4pt}1109\discretionary{/}{%
}{/}TVCG\hspace{.1pt}\discretionary{.}{%
}{.}\hspace{.4pt}2023\hspace{.1pt}\discretionary{.}{%
}{.}\hspace{.4pt}3327163}}}


\bibitem{neuronpedia}
Z.~Yu and S.~Ananiadou.
\newblock Neuronpedia: Interactive reference and tooling for analyzing neural networks.
\newblock https://www.neuronpedia.org/.
\newblock Neuronpedia, accessed July 2025.

\bibitem{yu-ananiadou-2024-neuron}
Z.~Yu and S.~Ananiadou.
\newblock Neuron-level knowledge attribution in large language models.
\newblock In Y.~Al-Onaizan, M.~Bansal, and Y.-N. Chen, eds., {\em EMNLP}, pp. 3267--3280. Association for Computational Linguistics, Miami, Nov. 2024. \href{https://doi.org/10.18653/v1/2024.emnlp-main.191}
{doi: {{%
10\hspace{.1pt}\discretionary{.}{%
}{.}\hspace{.4pt}18653\discretionary{/}{%
}{/}v1\discretionary{/}{%
}{/}2024\hspace{.1pt}\discretionary{.}{%
}{.}\hspace{.4pt}emnlp\discretionary{%
}{-}{-}main\hspace{.1pt}\discretionary{.}{%
}{.}\hspace{.4pt}191}}}


\bibitem{Human-in-the-loop}
Z.~Zhao, P.~Xu, C.~Scheidegger, and L.~Ren.
\newblock { Human-in-the-loop Extraction of Interpretable Concepts in Deep Learning Models }.
\newblock {\em {IEEE} Trans. Vis. Comput. Graph.}, 28(01):780--790, Jan. 2022. \href{https://doi.org/10.1109/TVCG.2021.3114837}
{doi: {{%
10\hspace{.1pt}\discretionary{.}{%
}{.}\hspace{.4pt}1109\discretionary{/}{%
}{/}TVCG\hspace{.1pt}\discretionary{.}{%
}{.}\hspace{.4pt}2021\hspace{.1pt}\discretionary{.}{%
}{.}\hspace{.4pt}3114837}}}


\end{thebibliography}

\end{document}